\documentclass[review]{elsarticle}

\usepackage{amsmath, amssymb, bm} 
\usepackage{enumitem}
\usepackage[utf8]{inputenc}
\usepackage{textcomp}
\usepackage{subcaption}
\usepackage{booktabs}
\usepackage{float}
\usepackage{makecell}
\usepackage [acronym]{glossaries}
\usepackage{comment}

\usepackage{silence}
\usepackage{etoolbox}

\usepackage{graphicx}
\usepackage{subcaption} % modern subfigures
\usepackage{upgreek}

\newtheorem{remark}{Remark}
\newacronym{rl}{RL}{Reinforcement Learning}
\newacronym{mpc}{MPC}{Model Predictive Control}
\newacronym{lpv}{LPV-\gls{mpc}}{Linear Parameter-Varying Model Predictive Control}
\newacronym{ndi}{NDI}{Nonlinear Dynamic Inversion}
\newacronym{indi}{INDI}{Incremental Nonlinear Dynamic Inversion}
\newacronym{lqr}{LQR}{Linear Quadratic Regulator}
\newacronym{lqg}{LQG}{Linear Quadratic Gaussian }
\newacronym{aoa}{AoA}{Effective Angle of Attack}
\newacronym{rms}{RMS}{Root-Mean-Square}
\newacronym{lpvmodel}{LPV}{Linear Parameter-Varying}

\newcommand{\mpcrl}{\gls{mpc}--\gls{rl}}
\newcommand{\tildebu}{\tilde{\bu}}
\newcommand{\bmf}{\bm{f}}
\newcommand{\bx}{\bm{x}}
\newcommand{\bz}{\bm{z}}
\newcommand{\bu}{\bm{u}}
\newcommand{\bv}{\bm{v}}
\newcommand{\ba}{\bm{a}}
\newcommand{\bd}{\bm{d}}
\newcommand{\bbeta}{\bm{\beta}}
\newcommand{\bxc}{\bm{x}^{\text{c}}}
\newcommand{\buc}{\bm{u}^{\text{c}}}
\newcommand{\bdc}{\bm{d}^{\text{c}}}
\newcommand{\vf}{V^{\text{f}}}
\newcommand{\pmpc}[1]{\pi^{\text{MPC}}\left( \bx(#1) \right)}
\newcommand{\xsafe}[1]{\mathcal{X}^{\text{safe}}\left(#1\right)}
\newcommand{\usafe}[1]{\mathcal{U}^{\text{safe}}\left(#1\right)}
\newcommand{\prl}{\pi^{\text{RL}}}
\newcommand{\umin}{\bu^{\min}}
\newcommand{\umax}{\bu^{\max}}
\newcommand{\xmin}{\bx^{\min}}
\newcommand{\xmax}{\bx^{\max}}
\newcommand{\vmin}{\bv^{\min}}
\newcommand{\vmax}{\bv^{\max}}
\newcommand{\amin}{\ba^{\min}}
\newcommand{\amax}{\ba^{\max}}
\newcommand{\lx}{\Bar{L}_{\bx}}
\newcommand{\lu}{\Bar{L}_{\bu}}

\newcommand{\xdep}{\bm{x}}
\newcommand\norm[1]{\left\lVert#1\right\rVert}

\newcommand{\xbar}{\Bar{\bm{x}}}
\newcommand{\xbarplus}{\Bar{\bm{x}}^{+}}
\newcommand{\ubar}{\Bar{\bm{u}}}
\newcommand{\barD}{\Bar{\mathcal{D}}}
\newcommand{\wbar}{w\left(\xbar\right)}
\newcommand{\repd}{\hat{\bd}}

\newcommand{\ndel}{\Bar{N}}

\newcommand{\barbdelta}{\Bar{\bm{\delta}}}
\newcommand{\bardelta}{\Bar{\delta}}
\newcommand{\lxi}[1]{L_{\bx}^{(#1)}(k)}
\newcommand{\luloc}{L_{\bu}(k)}
\newcommand{\wnorm}[1]{{\left\lVert#1\right\rVert}_W}
\newcommand{\unitvec}{\hat{\bm{i}}}
\newcommand{\OSpl}{{\Delta h}^{\max}(k)}
\newcommand{\OSaoa}{\alpha^{\text{eff},\max}(k)}
\newcommand{\STpl}{h^{\text{settle}}(k)}
\newcommand{\STaoa}{\alpha^{\text{eff},\text{settle}}(k)}

\newcommand{\pl}{h(k)}
\newcommand{\vpl}{v^{\text{h}}(k)}
\newcommand{\pitch}{\theta(k)}
\newcommand{\vpitch}{v^{\uptheta}(k)}

\newcommand{\Dfl}{\beta^{\text{f}}(k)}



\usepackage{graphicx}
\usepackage{amsmath}
\usepackage[version=4]{mhchem}
\usepackage{siunitx}
\usepackage{longtable,tabularx}
\begin{document}
\begin{frontmatter}

\title{MPC-Guided Safe Reinforcement Learning and Lipschitz-Based Filtering for Structured Nonlinear Systems} 
%:  An Application to Aeroelastic Wing Control}
% \author{P. Kostelac \footnote{MSc. Student, Faculty of Aerospace Engineering, Control and Simulation Division, Delft University of Technology.}}
% \affil{Delft University of Technology, Faculty of Aerospace Engineering, 2629 HS Delft, The Netherlands}
% \affil{Delft University of Technology, Faculty of Aerospace Engineering, Department of Control \& Operations, 2628 BX Delft, The Netherlands}

\author[1]{Patrick Kostelac\footnote{Patrick.Kostelac@gmail.com (corresponding author)}}%
\address[1]{Delft University of Technology, Department of Control and Operations, 2629 HS, Delft, The Netherlands}

\author[2]{Xuerui Wang\footnote{X.Wang-6@tudelft.nl}}%
\address[2]{Delft University of Technology, Department of Aerospace Structures and Materials, 2629 HS, Delft, The Netherlands}

\author[1,3]{Anahita Jamshidnejad\footnote{A.Jamshidnejad@tudelft.nl}}%
\address[3]{Delft University of Technology, Department of Intelligent Systems, 2628 XE, Delft, The Netherlands}

%\maketitle

\begin{abstract}
%
% \tableofcontents

% The evolution of aerospace engineering has led to aircraft with lightweight, flexible structures that improve fuel and aerodynamic performance. However, this flexibility increases sensitivity to gust disturbances, demanding control strategies that are robust, adaptive, and constraint-aware under real-time conditions. Linear Parameter-Varying Model Predictive Control (LPV-MPC) offers robustness and structured constraint handling. However, its effectiveness can be limited under turbulence and real-time constraints. Reinforcement Learning (RL), on the other hand, enables adaptive and lightweight control in uncertain, nonlinear systems, but lacks inherent safety guarantees. To address these limitations, a hybrid control framework that combines the safety and constraint-handling strengths of MPC with the adaptability of RL is proposed. During training, MPC defines safe control bounds that guide the RL agent, ensuring constraint satisfaction by design. This allows the agent to learn effective policies within a safety envelope, enabling lightweight, real-time execution at deployment. Simulation results demonstrate improved disturbance rejection, constraint satisfaction, and reduced computational demand compared to standalone approaches.
%
Modern engineering systems, such as autonomous vehicles, flexible robotics, and intelligent aerospace platforms, 
require controllers that are robust to uncertainties, adaptive to environmental changes, and safety-aware under real-time constraints. 
\gls{rl} offers powerful data-driven adaptability for systems with 
nonlinear dynamics that interact with uncertain environments. 
\gls{rl}, however, lacks built-in mechanisms for dynamic constraint satisfaction during exploration. 
\gls{mpc} offers structured constraint handling and robustness, 
but its reliance on accurate models and computationally demanding online optimization may pose significant challenges. 
This paper proposes an integrated \mpcrl\ framework that combines stability and safety guarantees of \gls{mpc} with the adaptability of \gls{rl}. 
During training, \gls{mpc} defines safe control bounds that guide the RL component 
and that enable constraint-aware policy learning. 
At deployment, the learned policy operates in real time with a lightweight safety filter based on Lipschitz continuity 
to ensure constraint satisfaction without heavy online optimizations. 
The approach, which is validated on a nonlinear aeroelastic wing system, 
demonstrates improved disturbance rejection, reduced actuator effort, and robust performance under turbulence. 
The architecture generalizes to other domains with structured nonlinearities and bounded disturbances, 
offering a scalable solution for safe artificial-intelligence-driven control in engineering applications.%
\end{abstract}


\end{frontmatter}

% \begin{multicols}{2}
% \newpage
\section{Introduction}
\glsresetall 

Modern control systems in engineering, e.g., autonomous vehicles, robotic manipulators, intelligent aerospace platforms, 
should operate reliably in uncertain and dynamically changing environments --  
often under strict real-time and safety constraints. 
These systems require controllers that handle high-dimensional state spaces and unmodeled dynamics, 
while ensuring safety, stability, and real-time constraint satisfaction \cite{robust-nonlinear-general-1, robust-nonlinear-general-2}. 
Traditional model-based methods offer structure and interpretability, but 
often struggle with model mismatch and scalability. 
Conversely, learning-based methods 
adapt to complex dynamics, but typically lack formal safety guarantees 
and may be data inefficient in constrained, safety-critical settings 
\cite{MPC-general-RL-changes-offroad, deepRL-book}.

\gls{mpc}, particularly \gls{lpv}, has gained traction for controlling systems 
with structured nonlinearities, due to its ability to enforce constraints, 
coordinate multi-variable inputs, and adapt to changing conditions in real time \cite{camacho2004mpcbook, D-MPC-LPV1, D-MPC-MLA}. 
In aerospace, \gls{lpv} has shown promise for flutter suppression and load alleviation 
\cite{D-MPC-MLA, D-MPC-LPV1, D-MPC-LPV2, D-MPC-LPVhe-su}. 
Many implementations rely on sufficiently accurate, reduced-order models \cite{D-MPC-wang}, 
but these are tedious to obtain and may be less reliable in presence of model uncertainties. 
Alternatively, preview-based gust estimations may be used \cite{D-MPC-LIDARpred}. 
\gls{lpv} often neglects actuator transport delay \cite{D-MPC-MaciejowskiEtAl-03}, 
and its performance depends heavily on linearization quality and parameter scheduling 
\cite{D-MPC-PereiraEtAl-19, D-MPC-MaciejowskiEtAl-03}. 
Moreover, repeated real-time optimization introduces significant computational demands, 
which limits the practicality of \gls{lpv} in onboard, safety-critical systems with fast dynamics \cite{Sanches2024-LPVsolver,darwich2022gust}.%\cite{MLA-phd}.%

\gls{rl} has emerged as a powerful data-driven framework for control in nonlinear, uncertain environments. 
It enables learning adaptive policies directly from interaction data \cite{RL-GLA-1, RL-SYSID-3}, 
making it attractive for systems where accurate modeling is difficult or infeasible. 
Prior studies have shown the potential 
of \gls{rl} in gust load alleviation, adaptive flight control, and fault-tolerant operation 
\cite{RL-adaptive-maneuvering-2, RL-GLA-2, RL-fault-tolerant-1}. 
\gls{rl} policies can be made lightweight at deployment, especially with simple architectures, 
such as Q-learning \cite{RL-stabilization-2}, 
which is adopted in this work for its simplicity and traceability. 
However, \gls{rl} methods typically lack built-in mechanisms for enforcing constraints 
and ensuring safety during training and deployment \cite{RL-SYSID-3, RL-adaptive-maneuvering-2}, 
limiting their direct applicability in safety-critical settings.%

To address these challenges, control architectures that integrate \gls{rl} 
with \gls{mpc} have gained attention. 
\gls{mpc} provides safety through predictive optimization and constraint enforcement within 
the online decision-making loop, 
while \gls{rl}, by learning from data, contributes adaptability and performance improvement  
in uncertain regimes. 
Together, \gls{mpc} and \gls{rl} enable safe, adaptive, efficient, and robust decision-making \cite{MPC-+-RL-for-uncertainties-or-disturbances}. 
This paper introduces a novel integrated \mpcrl\ framework, 
which differs from existing \mpcrl\ combinations by enabling parallel, 
yet cooperative roles: \gls{mpc} provides constraint-satisfying baseline actions, 
while \gls{rl} refines them for robustness under nonlinear disturbances. 
Unlike prior approaches where one method dominates (e.g., \gls{mpc} supervising \gls{rl} 
or \gls{rl} tuning \gls{mpc} parameters), our architecture achieves balanced 
integration with offline \gls{mpc}-guided safe policy learning and lightweight 
Lipschitz-based safety filtering at deployment.%

Emerging engineering systems increasingly exhibit high-order nonlinear dynamics, 
tight safety constraints, and sensitivity to external disturbances. 
One representative example is the control of modern aircraft with lightweight, aeroelastic wings ---   
structures designed to improve fuel efficiency and aerodynamic performance \cite{WINGFLEX1, WINGFLEX2}. 
These systems introduce complex control challenges due to strong coupling between structural 
deformation and unsteady aerodynamics. 
Aeroelastic phenomena, such as flutter, divergence, and control reversal 
pose serious risks to flight stability \cite{FLUTTER1}, 
especially under atmospheric turbulence. 
These scenarios demand precise real-time control despite actuator delays, 
model uncertainty, and strict input and state constraints \cite{FLUTTER1}. 
In this work, aeroelastic wing control is used as a case study 
to demonstrate the broader applicability of the proposed integrated 
\mpcrl\ architecture for safe, adaptive control in complex nonlinear systems.%

Beyond controller-level safety filters, formal verification methods provide deductive guarantees 
for hybrid and discrete-time systems. 
Classical barrier certificates certify inductive state invariants, 
but may be conservative or fail for some templates. 
Recent work introduces closure certificates (a functional analogue of transition invariants) 
that extend barrier-style arguments from state invariants to transition invariants. 
This allows for verification of safety, persistence (finite visits), and 
specifications in discrete-time systems \cite{Murali2026}. 
This perspective situates our deployment-time filtering within a broader certification landscape.
% 

% ---------------------------------------------------------------------------------------------------------------------------------------------------------

\subsection{Key Contributions \& Structure of the Paper}

While both \gls{mpc} and \gls{rl} have demonstrated success in various applications, 
their integration typically follows two dominant paradigms: 
First, \gls{mpc} is directly embedded into an \gls{rl} framework to 
enable constraint satisfaction and model-based reasoning. 
Alternatively, \gls{rl} is deployed to improve adaptability of \gls{mpc} 
by tuning its internal weights and constraints. 
To the best of our knowledge, no existing integrated \mpcrl\ 
architecture fully captures the complementary strengths 
of both methods in a unified, deployable framework. 
As such, one method often dominates and hinders specific strengths of the other one:   
In \gls{mpc}-guided \gls{rl}, inherent adaptability of \gls{rl} may be restrained, whereas 
in \gls{rl}-guided \gls{mpc} feasibility is put at risk. 
Such imbalance prevents the full potential of both methods from being realized in safety-critical, real-time settings.%

Alternatively, a unified architecture should enable \gls{mpc} and \gls{rl} to operate in parallel, yet cooperative roles. 
In such a framework, \gls{mpc} ensures feasibility and constraint satisfaction, 
while \gls{rl} enhances adaptability and robustness to nonlinear disturbances. 
This paper addresses the gap by proposing a new architecture, 
in which \gls{mpc} and \gls{rl} are connected in parallel, yet cooperative roles: 
\gls{mpc} provides constraint-satisfying baseline inputs, while \gls{rl} refines them  
to account for nonlinearities and to add robustness 
by considering control inputs in various disturbance-affected scenarios. 
The main contributions are:
\begin{itemize}
    \item 
    \textbf{A novel, unified integrated \mpcrl\ control architecture} 
    that combines offline policy generation with online reinforcement learning 
    to enable lightweight, constraint-aware policy learning for real-time control in nonlinear, delay-affected systems.
    \item 
    \textbf{A structured training strategy}, 
    in which \gls{mpc} defines state and disturbance-dependent control bounds under disturbances, 
    ensuring that all training actions remain within a guaranteed safe set.
    \item 
    \textbf{A deployment-time safety filter}, which leverages Lipschitz continuity to interpolate between certified safe actions. 
    It guarantees constraint satisfaction for states 
    near the training distribution, enabling lightweight, real-time execution without online optimization.
\end{itemize}
In order to ensure transparency and to reduce training complexity, 
the \gls{rl} component is implemented using tabular Q-learning, 
which provides a lightweight and interpretable alternative to deep \gls{rl}, 
while preserving long-term value estimation. 
Altogether, the proposed architecture offers a scalable and certifiable solution for safe \gls{rl} deployment in nonlinear, 
safety-critical systems, and generalizes beyond the 
given case study for aeroelastic control to domains, 
such as robotics, energy systems, and autonomous vehicles.%

\begin{figure}
    \centering
    \includegraphics[width=0.35\linewidth]{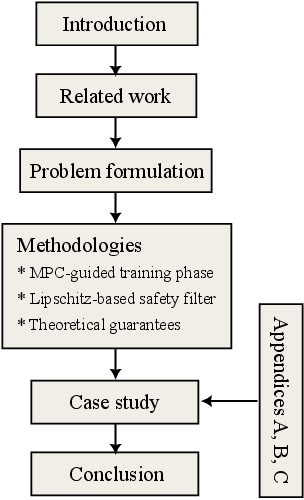}
    \caption{Roadmap of the paper.}
    \label{fig:roadmap}
\end{figure}

The rest of the paper is organized as follows: 
Section~\ref{sec:related_work} reviews related work on \gls{mpc}, \gls{rl}, and their integration strategies. 
Section~\ref{sec:problem_formulation} formulates the problem, including system class, control objectives, and operating assumptions. 
Section~\ref{sec:proposed_method} presents the proposed integrated \mpcrl\ framework 
and details the training and deployment phases, as well as theoretical safety guarantees. 
Section~\ref{sec:case_study} introduces the aeroelastic wing case study, 
reports the simulation results and performance comparisons across controllers, 
and discusses implications, limitations, and potential extensions. 
Finally, Section~\ref{sec:paper_conclusion} concludes the paper and outlines directions for future research. 
Figure~\ref{fig:roadmap} illustrates the structure and flow of these sections. 
A table including the frequently-used mathematical notations is also given in \ref{appendix:notations}.%

%------------------------------------------------------------------------------------------------------------------
\section{Related Work}
\label{sec:related_work}

%•	Gaps in current relevant control methods

%existing methods 1 (gaps so related work)
Classical linear controllers, e.g., \gls{lqr} \cite{LQR1}, \gls{lqg} \cite{LQR2}, and 
H-infinity control ($\mathcal{H}_\infty$) \cite{Hinf}, 
remain widely used due to their simplicity, computational efficiency, and ease of tuning. 
However, their reliance on accurate linearization and gain scheduling 
limits their effectiveness in nonlinear systems that are subject to external disturbances and actuation transport delay \cite{LQRDISTURBACE,Gainscheduling,LPVanalysis-gainscheduling}.% 

Nonlinear methods, e.g., \gls{ndi} \cite{NDI}, \gls{indi} \cite{INDI}, and backstepping \cite{Backstepping}, 
offer improved handling of nonlinear dynamics, but face challenges 
such as model dependency, sensitivity to internal instabilities, and computational complexity in high-order systems \cite{wang2019stability}.%

% lpv MPC (intro so challenges of MPC alone)
\gls{lpv}, which enables constraint enforcement and multi-variable coordination,  
has emerged as a structured approach for controlling 
nonlinear systems prone to uncertainties \cite{D-MPC-MLA}. 
\gls{lpv} has shown promise in enhancing performance and robustness \cite{D-MPC-LPV2,D-MPC-LPVhe-su}, 
but its reliance on reduced-order models, preview-based disturbance estimation, 
and neglecting delays, e.g., caused by system actuators, 
limit its reliability under realistic conditions \cite{D-MPC-wang, D-MPC-LIDARpred,D-MPC-PereiraEtAl-19}. 
Moreover, repeated real-time optimization -- inherent to various \gls{mpc} methods -- introduces computational burdens 
that challenge onboard deployment of \gls{lpv} \cite{Sanches2024-LPVsolver}.%

Several strategies have been proposed to integrate \gls{rl} and \gls{mpc} 
with the aim of balancing adaptability, safety, and computational complexity. 
Next, we briefly review three main categories of such integrated control methods.% 

% -------------------------------------------------------------------------------------------------

% •	Overview of prior MPC and RL integrations

\subsection{\gls{mpc} as Policy for \gls{rl}}
\label{sc:MPC_as_Policy_in_RL}

\gls{mpc} can serve as a structured policy within \gls{rl}, offering an   
alternative to neural-network-based policies. 
Instead of learning a direct state-to-action mapping, 
\gls{rl} components may rely on \gls{mpc} inputs generated by solving 
an optimization problem per time step, 
inherently enforcing system constraints and improving reliability 
\cite{overview-of-MPC-+-RL-survey-and-classification, SafeRL+Robust-MPC}. 
Given the current state $\bx(k)$ of the controlled system, \gls{mpc} solves for a sequence of control inputs 
$\tildebu^*(k) = \left\{\bu^*(k), \dots, \bu^*(k+N-1) \right\}$ that minimizes a finite-horizon cost 
function, such as:
\begin{equation}
\label{eq:MPC_cost_formulation}
\tildebu^*(k) = \arg\min_{\tildebu(k)} \left\{\sum_{\kappa = k}^{k+N-1} l\left(\bx(\kappa), \bu(\kappa)\right) + \vf\left(\bx(k+N)\right)\right\}
\end{equation}
subject to constraints $\bx(\kappa+1) = \bmf\left( \bx(\kappa) , \bu(\kappa) \right)$ 
and $\left( \bx(\kappa) , \bu(\kappa) \right) \in \mathcal{C}$, for $\kappa \in \{ k, \ldots, k+N-1 \}$. 
Here $N$ is the \gls{mpc} prediction horizon, 
$\bx(\kappa) \in \mathbb{R}^n$ and $\bu(\kappa) \in \mathbb{R}^m$ are 
the system state and control input at time step $\kappa$,
$l\left(\bx(\kappa), \bu(\kappa)\right)$ is the stage cost,
$\vf\left(\bx(k+N)\right)$ is the terminal cost,
$\bmf(\cdot,\cdot)$ is the prediction model of \gls{mpc}  
that describes dynamics of the controlled system, 
and $\mathcal{C}$ is the set of admissible state-input pairs. 
Although the optimization yields a full control sequence, 
only the first input is applied to the system, following the receding horizon principle \cite{camacho2004mpcbook}.
The policy of \gls{mpc} is therefore defined as $\pmpc{k} = \bu^*(k)$. 

\gls{mpc} may serve as a structured policy approximator 
within the \gls{rl} loop, whereas the \gls{rl} component 
learns to refine its parameters 
(e.g., cost weights, constraints, prediction horizon, prediction models) over time to 
enable data-efficient learning, while maintaining feasibility. 
This approach reduces the need for trial-and-error exploration and enhances interpretability
\cite{RL-based-on-RMPC,overview-of-RL-for-MPC-fundamentals-and-challenges, RL-based-on-MPC+stoch-policy-gradient}. 
\gls{mpc}-based policies can guide \gls{rl} exploration more effectively:   
For instance, variance-based exploration strategies use \gls{mpc} 
to generate feasible trajectories in high-uncertainty regions and to 
improve sample efficiency \cite{Variance-exploration-for-RL-MPC}. 
Robust \gls{mpc} may also act as a safety filter to allow \gls{rl} components 
to explore more freely, while ensuring constraint satisfaction \cite{multi-agent-RL-via-DMPC}. 
While promising, this integration introduces challenges, 
including increased computational complexity due to online optimization by \gls{mpc} 
and exploration demands by \gls{rl} \cite{RL-based-on-MPC+stoch-policy-gradient, overview-of-RL-for-MPC-fundamentals-and-challenges}. 
Nonetheless, the \gls{mpc}-as-policy paradigm offers a principled way to embed model-based safety 
into learning-based control, making it a compelling direction for robust, adaptive systems.%

% \begin{equation}
% \begin{aligned}
% \pi_{\text{MPC}}(s) = \arg\min_{u} & \sum_{k=0}^{N-1} l(x(k), u(k)) + V_f(x_N) \\
% \text{s.t.} \quad & x_{k+1} = f(x(k), u(k)), \quad (x(k), u(k)) \in \mathcal{C}
% \end{aligned}
% \end{equation}
% where $ V_f(x_N) $ is a terminal cost and $ \mathcal{C} $ represents constraints. RL can either treat this structure as fixed or adapt its parameters based on experience \cite{overview_of_MPC_+_RL_survey_and_classification}.

% ---------------------------------------------------------------------------------------------------
\subsection{RL Adjusting \gls{mpc} Parameters}

\gls{mpc} relies on fixed parameters (e.g., cost weights, prediction horizon, constraint sets)  
to solve constrained optimization problems. 
However, static tuning limits adaptability in dynamic environments and may cause infeasibility or degraded performance \cite{overview-of-MPC-+-RL-survey-and-classification}. 
\gls{rl} addresses this by adjusting parameters of \gls{mpc} in real time, 
reducing reliance on offline tuning and enhancing responsiveness 
\cite{RL-based-on-MPC+stoch-policy-gradient,RL-to-tune-NMPC,adaptive-parametrized-MPC-based-on-RL+synethis-framework,SYSID+RLMPC}

Based on \eqref{eq:MPC_cost_formulation}, consider the \gls{mpc} cost $
J(k; Q, R) = \sum_{\kappa = k}^{k + N-1} l\left(\bx(\kappa), \bu(\kappa); Q, R\right) + \vf\left(\bx(k + N)\right)
$
at time step $k$, where 
$Q \in \mathbb{R}^{n \times n}$ and $R \in \mathbb{R}^{m \times m}$ reflect  
trade-offs between state deviation and control effort. 
The \gls{rl} component may act as an outer-loop optimizer that learns to adjust 
$N$ or $Q$ and $R$ to improve robustness in changing conditions, 
based on the controller performance. 
Tabular Q-learning \cite{RL-stabilization-2} or policy gradient methods \cite{Lehmann2024PolicyGradients}
associate parameter configurations 
with expected returns, enabling adaptive tuning without manual intervention \cite{RL-based-on-real-time-NMPC}.%

Alternatively, the \gls{rl} component updates constraints of \gls{mpc}, i.e., 
$\bx(k) \in \mathcal{X}_{\boldsymbol{\theta}}$ and $\bu(k) \in \mathcal{U}_{\boldsymbol{\theta}}$,  
where $\boldsymbol{\theta}$ parameterizes the admissible state and input sets $\mathcal{X}$ and $\mathcal{U}$, respectively.  
This allows to relax or tighten constraints dynamically 
in order to avoid infeasibility under uncertainty, 
while preserving safety-critical margins \cite{safe-RL-using-NMPC-and-Policy-gradients}. 
Beyond these, the \gls{rl} component enables tuning of the terminal cost function $\vf(\cdot)$  
and internal parameters of the MPC prediction model function $f(\cdot,\cdot)$.%

To ensure stability, safe \gls{rl} methods apply Lyapunov-based constraints 
that enforce asymptotic stability, 
trust-region updates for limiting abrupt parameter changes,
and robust constraint adaptation for preserving feasibility under uncertainty 
\cite{safe-RL-using-NMPC-and-Policy-gradients, RL-within-MPC-framework}. 
These mechanisms allow \gls{rl} to refine \gls{mpc} parameters while maintaining safety, 
making \gls{rl}-tuned \gls{mpc} suitable for safety-critical applications, 
such as for autonomous flight, energy grid management, and industrial process control. 
\gls{rl}-tuned \gls{mpc}, however, increases computational demands and safety risks 
and faces challenges regarding sample efficiency and stability \cite{RL-for-improving-MPC-params}.%

% ---------------------------------------------------------------------------------------------------

\subsection{Alternative \gls{mpc} and \gls{rl} Combinations}
\label{sc:RL+MPC_other}

\gls{rl} can refine control inputs generated by \gls{mpc} to improve 
adaptability under model inaccuracies and disturbances, 
while \gls{mpc} ensures feasibility and constraint satisfaction. 
This integrated setup retains stability guarantees  by the \gls{mpc} framework, 
while enabling \gls{rl} to apply data-driven corrections \cite{MPC-general-RL-changes-offroad}. 
Examples are off-road driving, where \gls{rl} adjusts \gls{mpc}-generated acceleration to manage terrain variability; 
bipedal locomotion where \gls{rl}-refined \gls{mpc} fine-tunes foot placement, preserving stability \cite{MPC-general-RL-refines}; 
and traffic systems, where \gls{rl} modifies \gls{mpc}-generated signal timings or ramp metering 
to respond to congestion \cite{MPC-+-RL-for-uncertainties-or-disturbances, MPC+DRL-TRAFFIC}. 
These examples show that \gls{rl}-refined \gls{mpc} can enhance performance without compromising safety, 
although coordination between learning-based adaptation and optimization-based feasibility 
remains critical to avoid conflicting actions and to ensure stable integration 
\cite{RL+MPC-microgrids, RL-setpoints-MPC-control, RL-guidance-MPC-lowlevel}.%

\gls{mpc} may supervise \gls{rl} by filtering or over-riding unsafe actions 
to ensure constraint satisfaction during learning. 
In such architectures, \gls{rl} proposes control inputs, while 
\gls{mpc} validates or adjusts them before execution \cite{SafeRL+Robust-MPC}. 
This improves safety without direct incorporation of constraints 
within the \gls{rl} decision-making loop, though it may lead to over-conservatism.  
For instance, in microgrids, \gls{rl} handles high-level tasks, including 
source switching, whereas \gls{mpc} refines continuous power flow \cite{RL+MPC-microgrids}. 
Similarly, nonlinear \gls{mpc} has been deployed to filter unsafe actions of \gls{rl} in real time 
to enable safe exploration \cite{safe-RL-using-NMPC-and-Policy-gradients}. 
These show that \gls{mpc} supervision supports safe deployment of \gls{rl}, 
although challenges in exploration, computation, and robustness -- particularly under 
model mismatch -- persist \cite{SafeRL+Robust-MPC, RL+MPC-microgrids, safe-RL-using-NMPC-and-Policy-gradients}.%

In parallel \mpcrl\ architectures, both components operate independently  
each addressing a different control objective \cite{RL+MPC-microgrids, RL-setpoints-MPC-control, RL-guidance-MPC-lowlevel}.  
While \gls{mpc} ensures short-term feasibility, \gls{rl} handles long-term, data-driven planning. 
This modular setup improves flexibility and allows independent tuning, but 
different control policies should be coordinated to prevent conflicts or instability. 
Application examples include microgrid management and power distribution, 
where \gls{rl} handles scheduling or reserves, while \gls{mpc} ensures feasible power flow \cite{RL+MPC-microgrids, RL-setpoints-MPC-control}; 
in quadrotor navigation, \gls{rl} plans paths within unknown environments, 
and \gls{mpc} steers the vehicle dynamics and obstacle avoidance~\cite{RL-guidance-MPC-lowlevel}. 
In such applications parallel \mpcrl\ setups blend learning-based adaptability with optimization-based reliability, 
although careful synchronization is still required \cite{RL+MPC-microgrids, RL-setpoints-MPC-control, RL-guidance-MPC-lowlevel}.%

Beyond these combinations, recent work has emphasized the importance of deployment-time safety filtering in \mpcrl\ systems. 
Lipschitz-based filters can certify interpolated actions near trained states without requiring online optimization. 
This allows for real-time constraint satisfaction, even under actuator delays or unseen conditions \cite{safe-RL-using-NMPC-and-Policy-gradients}. 
This approach complements the supervisory role of \gls{mpc} and supports scalable, certifiable deployment of \gls{rl} in safety-critical domains. Existing approaches often suffer from high computational cost, 
poor online feasibility, or limited adaptability to nonlinear disturbances 
\cite{SafeRL+Robust-MPC, RL+MPC-microgrids, safe-RL-using-NMPC-and-Policy-gradients}. 
As a conservative alternative, tube-based tightening using robust positively invariant sets 
has been proposed for predictive learning controllers to guarantee constraint satisfaction 
under non-repetitive disturbances \cite{HE2024101436}.%

%------------------------------------------------------------------------------------------------------------------
\section{Problem Formulation}
\label{sec:problem_formulation}

This section defines the class of systems targeted by the proposed unified \mpcrl\ framework, 
outlines control objectives and constraints, and presents the operating assumptions under which the proposed method is applicable.%
\vspace{-3ex}

\subsection{System Class and Generalization}

The proposed integrated control framework addresses nonlinear systems 
with state vector $\bxc(t) \in \mathbb{R}^n$, control input vector $\buc(t) \in \mathbb{R}^m$, 
and subject to unknown bounded external disturbances $\bdc(t) \in \mathbb{R}^q$, 
real-time constraints, and safety-critical operation. Such systems are represented by the following continuous-time dynamics, where $t \in \mathbb{R}$:
\begin{align}
\label{eq:continuous_time_dynamic_equation}
\dot{\bxc}(t) = \bmf\left( \bxc(t) , \buc(t), \bdc(t) \right) 
\end{align}
with $\bmf: \mathbb{R}^n \times \mathbb{R}^m \times \mathbb{R}^q \to \mathbb{R}^n$ a smooth, generally nonlinear function. 
The control input $\buc(t)$ and disturbance $\bdc(t)$ may themselves depend on the state vector $\bxc(t)$, 
either explicitly or through structured channels, e.g., filtered actuator dynamics or feedback-based disturbance models.%

For implementation, we consider a discrete-time approximation of the continuous-time dynamics in  \eqref{eq:continuous_time_dynamic_equation}. 
Using the first-order Euler approximation, the one-step update is:
\begin{align}
\label{eq:dynamic_equation_euler}
\bx(k+1) \approx 
\bx(k)+T\,\bmf\left(\bx(k),\bu(k),\bd(k)\right) 
\end{align}
where $\bx(k)$, $\bu(k)$, and $\bd(k)$ denote the discrete-time state, input, and disturbance vectors 
at discrete time step $k$.

Due to practical latency effects (e.g., actuator lag, sensor latency, communication delays)  
for various systems --- where control inputs influence the dynamics 
through structured, monotonic relationships and local linearizations of the controlled system 
can be captured via parameter-varying models --- 
the dominant effect of $\bu(k)$ on the state 
vector appears at time step $k+2$. 
Examples of such systems include robotic manipulators, autonomous vehicles, underwater robots, flexible structures with delay-affected actuation. 
To reflect this dominant effect, a delay horizon of two time steps is chosen. 
Using a second-order Taylor expansion over a fixed sampling time $T$, the 
state update equation at time step $k + 2$ is approximated by: 
\vspace{-3ex}
\begin{align}
\label{eq:dynamic_equation}
\bx(k +2) \approx\ &\bx(k) + 2T\, \bmf\left( \bx(k), \bu(k), \bd(k) \right) +  \\
 T^2 \Big[ &\nabla_{\bx} \bmf\left( \bx(k), \bu(k), \bd(k) \right) + 
 \nabla_{\bu} \bmf\left( \bx(k), \bu(k), \bd(k) \right) \cdot \nabla_{\bx} \bu(k) + \nonumber \\
&  \nabla_{\bd} \bmf\left( \bx(k), \bu(k), \bd(k) \right) \cdot \nabla_{\bx} \bd(k) \Big] \cdot \bmf\left( \bx(k), \bu(k), \bd(k) \right) \nonumber
\end{align}
where $\nabla_{\bm{v}}\bmf(\cdot,\cdot,\cdot) $ is the Jacobian of 
function $\bmf(\cdot,\cdot,\cdot)$ with respect to vector $\bm{v}$. 
Terms involving order three or higher of $T$ have been neglected in the approximation.%  

% ------------------------------------------------------------

\subsection{Control Objectives and Constraints}

The primary control objective is disturbance rejection, i.e.,  
the controlled system should attenuate unknown external disturbances  
in real time, while maintaining safe, bounded state trajectories. 
In this work, deviations from the nominal state trajectory 
are attributed to unmeasured disturbances, 
and the controller is expected to restore the system to its nominal configuration. 
The \mpcrl\ framework can also be extended to accommodate commanded maneuvers  
(e.g., human inputs or reference tracking).%

The dynamics is subject to input and state constraints. 
Control input constraints (e.g., actuator saturation, rate limits) 
reflect physical bounds and transport delays. 
State constraints bound state-dependent variables 
(e.g., displacement, velocity) to ensure structural integrity and mission safety. 
We use box constraints for simplicity and computational efficiency, 
but the framework may be extended to support alternative (e.g., polytopic, nonlinear)  constraint sets.  
This requires additional theoretical guarantees, 
e.g., invariant set analysis or robust feasibility conditions, to ensure safety under all admissible operating conditions.%

 % -----------------------------------------------------------------------------------

\subsection{Operating Conditions and Assumptions}

We consider the following conditions and assumptions:
\begin{enumerate}[label=\textbf{A\arabic*}]
\item \textbf{Smoothness:} \label{ass:smoothness}
The controlled dynamics $\bmf(\cdot,\cdot,\cdot)$  is 
smooth and locally Lipschitz in all arguments. 
\item \textbf{Boundedness \& admissibility:} \label{ass:bounded_d}
Disturbances are bounded within set $\mathcal{W}$. 
State and control input belong to admissible sets 
$\mathcal{X}$ and $\mathcal{U}$, respectively. 
\item \textbf{Delay-aware actuation:} \label{ass:delay_aware}
The system exhibits structured delays, such as actuator lag or sensor latency, 
which are captured via a fixed rollout horizon.
\item \textbf{Monotonicity:} \label{ass:monotone}
Control inputs influence the system through monotonic or affine channels, 
allowing interpolation between known safe actions.
\item \textbf{Training-time model access:} \label{ass:training_access}
During training, the \gls{mpc} component leverages the full system model and 
representative disturbances to simulate diverse scenarios and 
to compute safe control bounds, which guide policy learning of the \gls{rl} component.%
\item \textbf{Deployment-time uncertainty:} \label{ass:deployment_uncertainty}
At deployment, the typical dependency on disturbance preview measurements (e.g., 
using LiDAR-based predictions) is removed, and no disturbance model is available to the controller. 
Unlike many existing \mpcrl\ approaches that rely on online optimization or disturbance forecasting, 
our framework ensures safety through a learned policy combined with a Lipschitz-based safety filter. 
This design enables lightweight, certifiable real-time execution without requiring heavy computation or model access 
(See, e.g., \cite{HE2024101436}, which represents a more conservative certification pathway 
aligned with hybrid systems practice).%
\end{enumerate}

% -----------------------------------------------------------------------------------------------------------
\section{Proposed Method}
\label{sec:proposed_method}

%•	Overview of the integrated RL-MPC architecture (Include Figures 1 and 2, or updated versions of them that also reflect the feedback you got from your defense)

\begin{figure*}
    \centering
    \includegraphics[width=0.9\textwidth]{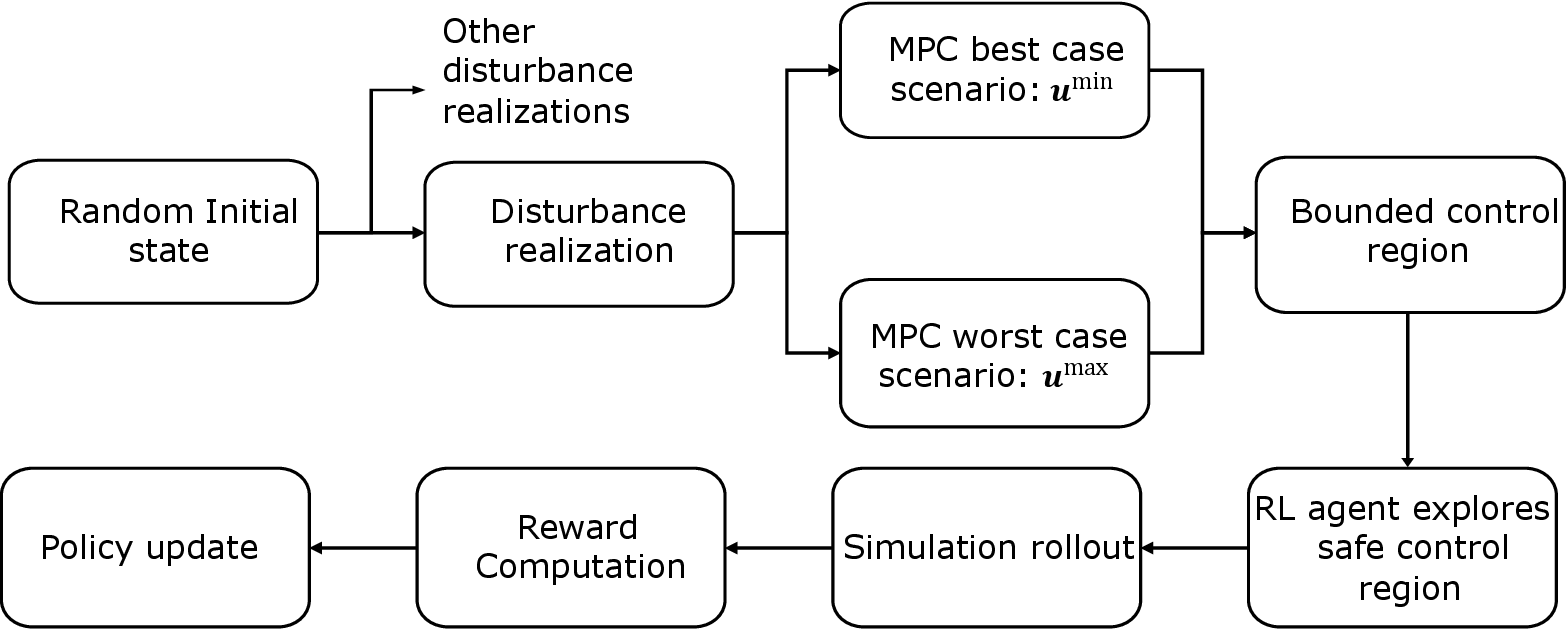}
    \caption{Overview of the integrated \mpcrl\  framework during the training phase.}
    % The illustrated setup corresponds to the control of an aeroelastic wing, as represented in the case study. 
    % Initial states represent wing configurations and disturbances are vertical gusts generated using a Dryden model. 
    % While the structure is general, the figure reflects this specific application.}
    \label{fig:training}

\vspace{5ex}

    \centering
    \includegraphics[width=0.9\textwidth]{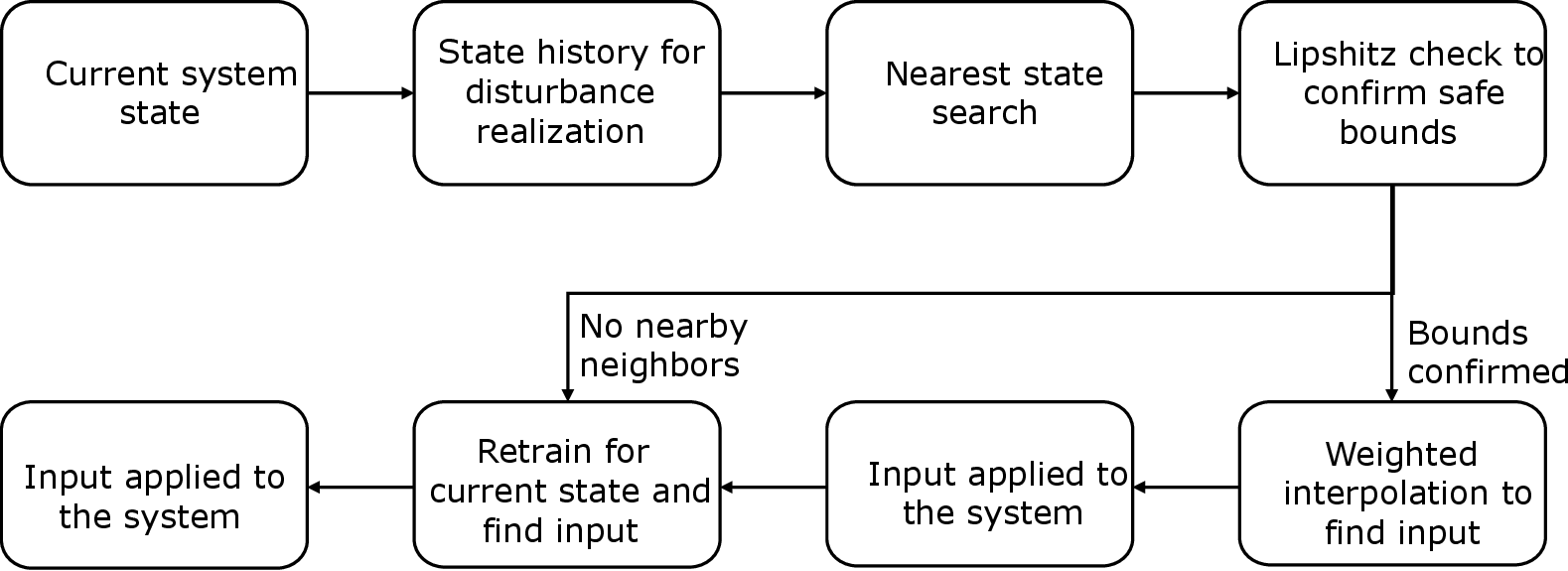}
    \caption{Overview of the integrated \mpcrl\ framework during the deployment phase.} 
    % The illustrated setup corresponds to the control of an aeroelastic wing, as represented in the case study. 
    % Interpolated control actions are selected based on previously verified safe trajectories under gust disturbances. 
    % While the structure is general, the figure reflects this specific application.}
    \label{fig:deployment}
\end{figure*}
% placing it above to appear in the right place

%  This section presents the control methodology proposed for safe learning and control of a flexible aeroelastic aircraft model. The approach hybridizes an LPV-based MPC to generate safe bounds on control actions and a RL policy to select feasible control inputs within those bounds. To validate this architecture, both the supporting numerical analysis \ref{subsec:Num_Val} and relevant theoretical guarantees \ref{subsec:Theoretical-Guaranteed} are developed. First a high level overview of the controller is shown with the training phase in Figure~\ref{fig:controller_architecture} and the deployment phase in Figure~\ref{fig:controller_architecture2}.

This section presents the proposed integrated \mpcrl\ framework, 
which integrates \gls{mpc} %{\color{red}(more specifically \gls{lpv})} 
and \gls{rl} in a parallel, cooperative structure. 
The framework ensures constraint satisfaction and disturbance rejection 
in nonlinear systems with bounded disturbances and delay-affected dynamics, 
as described in \eqref{eq:continuous_time_dynamic_equation}.%

The framework operates over a fixed delay order $\ell$, i.e., 
the number of discrete time steps 
required for control inputs to influence the system states. 
While the architecture supports arbitrary values for $\ell$, we focus on systems with a second-order delay, i.e., $\ell = 2$, 
which is sufficient to capture dominant nonlinear behaviors (e.g.,   
actuator lag, indirect coupling effects, sensor latency) while balancing model fidelity and computational tractability.  
Higher-order terms can often be neglected when the sampling time is sufficiently small and the system dynamics is well-behaved. 
% This simplification is justified when the third-order contributions are significantly smaller than the second-order terms, 
% and when the system exhibits bounded curvature and moderate state rates. 
This choice is further supported by numerical analyses for aeroelastic wing  
presented in Section~\ref{sec:case_study}, 
where the relative contribution of third-order terms is less than $2\%$ across all state components for 
sampling time $T = 10^{-3} \text{s}$.%

% ----------------------------------------------------------------------------------------------------
\subsection{Training Phase: \gls{mpc}--Guided Policy Learning}

Unlike conventional \gls{rl} approaches that rely on random exploration, the proposed framework employs 
structured sampling of initial states to ensure full coverage of the operational envelope. 
These states are uniformly distributed across relevant dimensions 
(e.g., displacement, orientation, velocity) to avoid clustering and under-representation of some state sub-regions. 
This promotes generalization of the learned policy across the entire feasible state space.%

For each initial state, a set of representative disturbance realizations is considered. 
Since the training phase faces no manually induced or unmodeled inputs, as per Assumption~\ref{ass:training_access}, 
the evolution of the nonlinear system is driven solely by external disturbances. 
% Depending on the state characteristics, between 5 and 15 disturbance realizations are used per wing state.
The number of realizations per state is selected to balance robustness across the operational envelope and computational tractability.% 

This setup allows to systematically associate plausible disturbance profiles with each state 
and eliminates the need for traditional learning curves. 
In standard \gls{rl}, repeated sampling and averaging are required to handle stochastic transitions. 
In contrast, the proposed training treats each state-disturbance pair as a deterministic micro-environment. 
While the ensemble of disturbances ensures robustness, fixing a disturbance realization for a given state yields a unique system response and state-action trajectory. 
Accordingly, the \gls{rl} component requires only a single interaction to extract the optimal control input and reward.%

The \gls{mpc} component computes the  constraint-satisfying bounded 
state set $\xsafe{\bx(k),\bd(k)} \subseteq \mathcal{X}$ 
and action set $\usafe{\bx(k),\bd(k)} \subseteq \mathcal{U}$ 
for each state–disturbance pair $\left( \bx(k), \bd(k) \right)$. 
The \gls{rl} component then learns a policy $\prl: \mathcal{X} \rightarrow \usafe{\bx(k),\bd(k)}$ within 
these bounds, ensuring that all training actions remain safe by construction. 
% Here, $\mathcal{X}$ and $\mathcal{U}$ denote the admissible sets for the state and control input, respectively.%

Figure~\ref{fig:training} summarizes the training process of the proposed \mpcrl\ control framework.  
It illustrates how the \gls{rl} component is trained to safely operate the nonlinear system within a bounded control region derived via the \gls{mpc} component. 
Each step --- from selection of the initial state  to disturbance sampling, 
action selection, and reward evaluation --- is structured to ensure both safety and coverage of the operational envelope.%

\begin{comment}
{\color{red}
\begin{remark}
    While Figure~\ref{fig:training} illustrates a specific instantiation of the training architecture for the aeroelastic wing case study, 
    the underlying structure, including safe action set generation, disturbance-aware sampling, and constraint-guided learning, 
    is applicable to a wide range of nonlinear systems with structured dynamics and bounded disturbances. 
    In Figure~\ref{fig:training}, initial states correspond to different aeroelastic wing configurations, 
    and the disturbances represent vertical gust profiles. 
    Two \gls{lpv} components compute lower and upper safe control bounds for each state–disturbance pair, 
    and the \gls{rl} component learns a policy within these bounds.%
\end{remark}
}
\end{comment}
% ------------------------------------------------------------------------------------------------------------------

\subsection{Deployment Phase: Safe Real-Time Execution}

During deployment, the system may encounter states that were not explicitly visited during training. 
To handle such cases, the controller interpolates between nearby trained states weighted based on proximity and constrained by operational limits of the system. This ensures that the resulting control input remains within the safe action bounds that have been established during training.%

This deployment strategy enables the control framework to generalize safely to new states, without requiring online optimization or access to the system model. 
The learned policy, combined with a Lipschitz-based safety filter, ensures that interpolated actions yield constraint-satisfying trajectories. 
This approach supports lightweight, certifiable execution suitable for safety-critical systems with fast dynamics and limited computational resources.%

In rare cases where the current state lies outside the interpolative regime (e.g., during extreme disturbance events), 
the integrated control framework triggers a fallback mechanism.  
This involves retraining the \gls{rl} component on the current state to compute a valid control action. 
While this incurs additional computation, the offline training phase is designed to densely cover the operational envelope, and 
to make such situations highly unlikely in practice.%

Figure~\ref{fig:deployment} illustrates the deployment phase of the proposed integrated \mpcrl\ control framework. 
During real-time operation, the controller deploys the policy learned 
per state-disturbance pair during training. 
To ensure safety and generalization, a runtime interpolation strategy is employed 
that maintains the architectural goals of scalability, robustness, and constraint satisfaction, 
without relying on heavy online computation.%

\begin{comment}
{\color{red}
\begin{remark}
    While Figure~\ref{fig:deployment} reflects the aeroelastic wing control --- 
    our case study example ---  
    the underlying principles of safe interpolation, constraint filtering, and fallback mechanisms are broadly applicable to other domains with structured nonlinearities and bounded uncertainties.%
\end{remark}
}
\end{comment}
% ------------------------------------------------------------------------------------------------------------------

\subsection{Theoretical Foundations for Constraint Satisfaction}
\label{subsec:Theoretical-Guaranteed}

A theoretical foundation is provided to guarantee 
constraint satisfaction during training and deployment 
of the \gls{rl} component. 
Formal guarantees are established to ensure that any control input within 
a bounded, verified safe set yields a 
state trajectory that satisfies the constraints. 
Let $\umin$ and $\umax$ be the element-wise lower and upper bounds 
for control input determined by the \gls{mpc} component during training. 
They define a verified safe action set, which satisfies the constraints. 
The analysis focuses on second-order state update equations, as given in \eqref{eq:dynamic_equation}, 
and shows that under certain structural assumptions, 
including dynamics monotonicity and smooth actuation response, the 
two-step resulting states 
remain bounded between states corresponding to control input bounds 
$\umin$ and $\umax$.% 

% ------------------------------------------------------------------------------------------------------------------

\subsubsection{Monotonicity and Boundedness of the Full State Update}

Here we show that for a nonlinear dynamical system with smooth control-affine structure 
governed by a second-order update equation, 
applying any intermediate control input 
$\umin\leq \bu(k) \leq \umax$ (element-wise) results in 
two-step updated states $\bx(\kappa)$, for $\kappa = k + 1, k + 2$, that remain bounded 
between $\xmin(\kappa)$  and $\xmax(\kappa)$, resulting from implementing control inputs $\umin$ and $\umax$, respectively.  
Such boundedness enables to infer constraint satisfaction by verifying only the bounding trajectories:  
Since $\xmin(\kappa)$  and $\xmax(\kappa)$ are generated by \gls{mpc} inputs, 
they remain within admissible state bounds, 
and due to the boundedness property, so does any intermediate state trajectory. 
This property is critical, as it forms the theoretical foundation for certifiable 
safe action selection by the \gls{rl} component.% 

To formally guarantee that the updated states lie between the trajectories generated by $ \umin $ and $ \umax $, certain conditions should hold. 
Specifically, the mapping $ \bm{\nu} \mapsto \bmf(\bx(k), \bm{\nu}, \bd(k)) $ 
should be monotonic for each individual state component 
to ensure that the effect of control inputs does not reverse direction (Assumption~\ref{ass:monotone}). 
Additionally, the dynamic update structure captured by the second-order Taylor expansion  
should preserve both the sign and relative magnitude of variations in $ \bmf\left(\bx(k), \bm{\nu}, \bd(k)\right) $ across the state evolution. 
Under these assumptions, bounding $ \bmf\left(\bx(k), \bm{\nu}, \bd(k)\right) $ between its values at the control input bounds  
is sufficient to bound the future state. However, this implication cannot be relied upon unless both monotonicity of the control-to-dynamics mapping 
and consistency in the second-order update structure are satisfied. 
Specifically, if the mapping  $\bm{\nu} \mapsto \bmf(\bx(k), \bm{\nu}, \bd(k))$ is not monotonic, 
or if the second-order expansion includes nonlinear interactions and couplings 
that distort the ordering of state updates, such as those mediated by actuator dynamics or disturbance-dependent feedback, 
then intermediate control inputs may yield trajectories that fall outside the bounds defined by $\umin$ and $\umax$.% 

Therefore, when monotonicity of the mapping $\bm{\nu} \mapsto \bmf\left(\bx(k), \bm{\nu}, \bd(k)\right)$ or consistency 
in the second-order update structure cannot be guaranteed globally, a more granular approach is required. 
Specifically, each state component should be analyzed individually to rigorously establish whether the state update 
resulting from an intermediate control input lies between those generated by the control bounds $\umin $ and $ \umax$. 
This component-wise analysis enables the certification of safe control actions, 
even in systems with complex nonlinearities, 
actuator dynamics that mediate the effect of control inputs on the state, 
or disturbance-dependent feedback mechanisms 
that may violate global monotonicity or ordering preservation.%

\begin{remark}{\textbf{Limitations of Monotonicity Assumption \ref{ass:monotone}:}} 
The theoretical guarantees in this section rely on Assumption~\ref{ass:monotone}, which posits that control inputs influence the system through monotonic or affine channels. 
While this holds for many systems with structured actuation (e.g., low-pass filtered actuators or single-input channels), it may not generalize to systems with strong nonlinearities, non-affine input effects, or coupled multi-input dynamics. 
In such cases, the monotonicity of the mapping $\bm{\nu} \mapsto \bmf\left(\bx(k), \bm{\nu}, \bd(k)\right)$ may not hold globally, 
and the bounding arguments presented here may no longer be valid. 
For these systems, more conservative safety certification methods, 
such as local reachability analysis or robust invariant set computation  
may be required to ensure constraint satisfaction (see, e.g., \cite{Murali2026,Shim2025}).%
\end{remark}

% ------------------------------------------------------------------------------------------------------------------

\subsubsection{Bounding Analysis for Second-Order State Components}
\label{sec:bounding_2nd_order_states}

This section establishes whether the second-order updates of individual state components 
$x^{(i)}(\kappa)$, for $i\in \{1,\ldots,n\}$ and $\kappa = k+1, k+ 2$, 
remain bounded between states resulting from control input bounds $\umin$ and $\umax$. 
Establishing this property ensures that applying any intermediate control input 
$\umin \leq \bu(k) \leq \umax$ yields a state trajectory 
within a known safe region, assuming that the bounding trajectories are verified to satisfy all constraints. 
The analysis is carried out component-wise and supports the certification of safe control actions 
under the structural assumptions introduced earlier, including monotonicity of control influence and smooth actuator dynamics.%

%----------------------------------------------------
\paragraph{Step 1. State Dynamics:}  
Let the full system state $\bx(k) =\left[ \bz^\top(k), \bv^\top(k) \right]^\top$ be composed of displacement state vector $\bz(k) \in \mathbb{R}^{n_1}$ 
and velocity state vector $\bv(k) \in \mathbb{R}^{n_2}$, with $n_1 + n_2 = n$. 
Here, $v_i(k)$ denotes the velocity corresponding to the displacement state $z_i(k)$ at time step $k$,  
for $ i\in \left\{1, \dots, n_1 \right\}$. 
The acceleration components $a_i(k)$ are not treated as 
separate state variables, but are assumed to be computable from the system dynamics and actuator states, 
as it is detailed in this section.%

Displacement components $z_i(k)$ evolve according to a   
standard kinematics second-order discrete-time state update 
based on the second-order Taylor expansion in \eqref{eq:dynamic_equation}: 
\begin{align}
\label{eq:component_displacement_state_update}
z_i(k+2) = z_i(k) + T v_i(k) + \dfrac{T^2}{2} a_i(k)
\end{align}
This relationship simplifies the update for each displacement state dimension by assuming that the 
corresponding velocity $v_i(k)$ and acceleration $a_i(k)$ are either known or computable. 
We assume that the only component that introduces the effect of control inputs $u_j(k)$ for $j = 1, \ldots, m$ to this dynamics is the acceleration $a_i(k)$. 
For velocity component $v_{\ell}(k)$, with $\ell = 1, \ldots, n_2$, the update follows:
\begin{align}
\label{eq:component_velocity_state_update}
v_{\ell}(k+2) =  v_{\ell}(k) + T a_{\ell}(k) 
\end{align}

% --------------------------------------------------------
\paragraph{Step 2. Acceleration Formulation:}  
The system is steered through actuators with state variable $\bbeta(k) \in \mathbb{R}^m$. 
Actuators mediate the effect of control inputs on the state $\bx(k)$. 
In systems where each control input component $u_j(k)$ corresponds to a distinct actuator state, it is natural to assume that $\bu(k)$ and $\bbeta(k)$ are of the same dimension. 
Each actuator state component $\beta_j(k)$, with $j = 1, \ldots, m$,  
introduces delay or filtering effects and evolves according to the following first-order low-pass filter dynamics\footnote{This discrete-time formulation 
corresponds to a forward Euler approximation of a continuous-time first-order actuator model.}, with $\lambda_j$ the actuator gain:
\begin{equation}
    \label{eq:Actuator_dynamics}
    \beta_j(k + 1) =  \left(1 - T \lambda_j\right) \beta_j(k) + T \lambda_j u_j(k)
\end{equation}
This formulation  ensures that,  over short time intervals, 
$\bbeta(k)$ is a smooth 
and monotonic function of control input $\bu(k)$. 
Moreover, since each actuator state component $\beta_j(k)$ evolves as a convex combination 
of its current value and current control input component $u_j(k)$, 
the actuator state vector $\bbeta(k)$ remains bounded within trajectories generated by 
control input bounds $\umin$ and $\umax$.%

We assume that the acceleration term $a_l(k)$, with $l = 1, \ldots, n$, for each state component $x_l(k)$ --- 
which may be a displacement or velocity component --- is an affine function of a control-influenced intermediate quantity, which depends on the control input via a smooth actuator state $\beta(k)$. 
This acceleration is modeled by: 
\begin{align}
\label{eq:acceleration}
a_l(k) = \alpha_l \left(\psi_l(\bbeta(k)) +  \phi_l\left(\bx(k), \bv(k)\right)\right) + \gamma_l(k)
\end{align}
where $\alpha_l \in \mathbb{R}$ is a scalar gain and $\gamma_l(k) \in \mathbb{R}$ is a scalar offset 
that may depend on the current state and velocity, but is independent of the control input. 
Here, $\psi_l(\cdot)$ captures control dependency of the acceleration via the actuator state vector $\bbeta(k)$, 
and $\phi_l(\cdot,\cdot)$ captures the state dependency.%

% -------------------------------------
\paragraph{Step 3. Monotonicity of Control Influence:}  
Since $\psi_i\left(\bbeta(k)\right)$ is monotonic in $\bbeta(k)$ 
(as per Assumption~\ref{ass:monotone}) and 
$\bbeta(k)$ is monotonic in $\bu(k)$ (as per \eqref{eq:Actuator_dynamics}), 
the acceleration vector $\ba(k)$ given by \eqref{eq:acceleration} is monotonic in $\bu(k)$. 
Moreover, as reasoned earlier, $\bbeta(k)$ given by \eqref{eq:Actuator_dynamics} remains bounded, 
which ensures that all control-dependent components of $\ba(k)$ remain within a certified safe envelope. 
In other words, the acceleration vector $\ba(k)$ satisfies: 
\begin{align}
\label{eq:bounded_acc}
\amin(k) \leq \ba(k) \leq \amax(k)
\end{align}
where $\amin(k)$ and $\amax(k)$ denote the acceleration vectors corresponding to injecting 
control input bounds $\umin$ and $\umax$ into the system at time step $k$.% 

Given \eqref{eq:component_velocity_state_update} and \eqref{eq:bounded_acc}, 
the updated velocity state $\bv(k+2)$ is also monotonic in $\bu(k)$, 
and satisfies $\vmin(k+2) \leq \bv(k+2) \leq \vmax(k+2)$, 
where $\vmin(k+2)$ and $\vmax(k+2)$ are the velocities at time step $k+2$ 
under, respectively, $\umin$ and $\umax$ applied at time step $k$. 
With both $v_i(k)$ and $a_i(k)$ bounded, based on \eqref{eq:component_displacement_state_update}, 
the two-step updated displacement state $\bz(k+2)$ is also bounded between its values under 
$\umin$ and $\umax$. Therefore, the full two-step state update satisfies the following condition:
\begin{align}
\xmin(k+2) \leq \bx(k + 2) \leq \xmax (k+2)
\end{align}

While the second-order expansion in \eqref{eq:dynamic_equation} is used to certify safety at 
time step $k+2$, the one-step state 
update given by \eqref{eq:dynamic_equation_euler} 
shows that $\bx(k+1)$ is also affected by $\bu(k)$. 
Moreover, due to actuator dynamics \eqref{eq:Actuator_dynamics}, 
the state $\bx(k+1)$ depends not only on $u_k$, but also on $u_{k-1}$. 
Under the monotonicity assumptions and the bounded trajectories generated by $\umin$ and $\umax$, 
the one-step updated state vector $\bx(k+1)$ remains bounded between 
$\xmin(k+1)$ and $\xmax(k+1)$ for any intermediate inputs.% 

Therefore, both intermediate states within the delay horizon 
satisfy the constraints, provided the bounding trajectories do.%

% ------------------------------------------------------------------------------------------------------------------
\subsection{Deployment-Time Safety Guarantees}
\label{sec:deployment_time_guarantees}

During real-world deployment, the system may encounter states that were not explicitly visited during training. 
To ensure continued constraint satisfaction, we introduce a runtime safety filter 
that certifies interpolated control actions based on locally verified data.%

This filter leverages the Lipschitz continuity of the system dynamics 
and a database of previously verified safe transitions.  
By interpolating control inputs from nearby training samples 
and bounding the deviation of the resulting state using Lipschitz constants, 
the method guarantees that the interpolated action yields a constraint-satisfying state update. 
This enables lightweight, simulation-free safety certification at runtime 
without requiring online optimization, model access, or multi-step forward simulation, 
even in the presence of actuator delays.%
% ------------------------------------------------------------------------------------------------------------------

\subsubsection{Local Database Construction for Safe Interpolation}

Let the current system state at deployment be denoted by $\xdep(k)$, 
which may not coincide with any state encountered during training 
and stored in the overall dataset $\barD$. 
To construct a safe control input, we interpolate from a local database 
$\mathcal{D}\left( \xdep(k) \right) \subseteq \barD$, 
composed of verified transitions within a neighborhood $\mathcal{N}\left( \xdep(k) \right)$ 
around $\xdep(k)$ (e.g., defined by a distance threshold or $k$-nearest neighbors): 
\begin{align}
\label{eq:local_database_construct}
\mathcal{D}\left( \xdep(k) \right) = \Big\{ \left(\xbar,\ \ubar,\ \xbarplus\right) \in \barD\, \Big| \,
\xbar \in \mathcal{N}\left( \xdep(k) \right) \Big\}
\end{align}
Each tuple $\left(\xbar, \ubar, \xbarplus\right)$ corresponds to a known safe one-step transition 
generated during training by simulating the system dynamics from an initial state $\xbar$ under a representative disturbance and applying control input $\ubar$. 
The resulting next state $\xbarplus$ is verified to satisfy all constraints, i.e., $\xbarplus \in \mathcal{X}^{\text{safe}}$. 
These verified transitions are used to construct and certify safe control inputs at deployment.%

Note that the disturbance is not included in these tuples, 
as it is assumed to be unmeasured at deployment 
(cf. Assumption~\ref{ass:deployment_uncertainty}). 
Instead, the database entries are constructed using representative disturbance realizations during training, 
and only transitions that satisfy all constraints under the applied disturbance -- 
thus guaranteeing robustness -- are retained. 
These verified transitions are then used to construct and certify safe control inputs at deployment.%

% -------------------------------------------------------------------------

\subsubsection{Control Synthesis and Lipschitz Deviation Bounds}

Given a query state $\xdep(k)$, the interpolated control input is computed via:
\begin{align}
\bu^* (k)= \sum_{\left(\xbar,\ \ubar,\ \xbarplus\right) \in \mathcal{D}\left( \xdep(k) \right)} \wbar \cdot \ubar
\end{align}
where the weights satisfy:
\begin{subequations}
\begin{align}
& \wbar \propto \dfrac{1}{\norm{\xdep(k) - \xbar} + \epsilon}\\
& \sum_{\left(\xbar,\ \ubar,\ \xbarplus\right) \in \mathcal{D}\left( \xdep(k) \right)} \wbar = 1
\end{align} 
\end{subequations}
with $\epsilon > 0$ a small regularization constant that ensures numerical stability.%

The objective is to show that, under mild regularity assumptions on the dynamics, the one-step state resulting from the interpolated input $\bu^*(k)$ remains close to the verified next-step state vector $\xbarplus$.  
This deviation is bounded using a Lipschitz expression, and if the bound remains within a predefined safety margin to the constraint boundary, 
the interpolated state is guaranteed to be safe.%

This filtering scheme is applied at every time step during deployment 
to provide formal, simulation-free safety certification. 
Unlike rollout-based approaches that rely on forward simulation to assess safety, 
this method guarantees safety through local deviation bounds  
and avoids multi-step prediction, even in the presence of actuator delays. 
This enables lightweight, certifiable deployment of learned policies in safety-critical systems with fast dynamics and limited computational resources.%

\paragraph{Global safety guarantee:} 
To certify the safety of the interpolated control input $\bu^*(k)$, 
a global Lipschitz-based bound on the deviation of the next-step state 
is established. 

Consider one-step state update: 
$\xdep(k+1) = \xdep(k) + T\;\bmf\left(\xdep(k), \bu^*(k), \repd \right)$, 
where $\repd \in \mathcal{W}$ denotes an arbitrary disturbance within 
the bounded set defined in Assumption~\ref{ass:bounded_d}. 
This notation allows to construct Lipschitz-based bounds, 
while the actual disturbance at deployment remains unknown (as per 
Assumption~\ref{ass:deployment_uncertainty}). 
The interpolated state is compared to the verified next-step state 
$\xbarplus = \xbar + T\; \bmf\left(\xbar, \ubar, \repd\right)$.
Assuming that  
$\bmf \left(\cdot, \cdot, \cdot\right)$ is Lipschitz continuous in all arguments 
(Assumption~\ref{ass:smoothness}), 
there exist constants $\lx$ and $\lu$ such that:
\begin{align}
\norm{\bmf\left(\xdep(k), \bu^*(k), \repd \right) - \bmf\left(\xbar, \ubar, \repd \right)} \leq \lx \norm{ \xdep(k) - \xbar} + \lu \norm{\bu^*(k) - \ubar}
\end{align}

Applying the triangle inequality, the deviation between $\xdep(k+1)$ and $\xbarplus$ satisfies:
\begin{align}
\label{eq:deviation_bound}
\norm{\xdep(k+1) - \xbarplus} \leq \left(1 + T \lx \right) \norm{\xdep(k) - \xbar} + T \lu \norm{\bu^*(k) - \ubar} 
\end{align}
This bound depends on the proximity of the current state to its neighbor states 
stored in $\mathcal{D}\left(\xdep(k)\right)$ 
and the similarity of the interpolated control input to the stored safe actions. 
It provides a \emph{global norm-based guarantee}, i.e.,  
the interpolated next-step state remains close to a verified safe state 
under bounded Lipschitz constants. 

Practical safety constraints may typically be defined per state component $x_i(\cdot)$. 
To address this, we refine the global bound into component-wise safety checks next.%

\begin{remark}{\textbf{Interpolation guarantees under delays:}} 
    Suppose that, due to actuator delays, the control input $\bu^*(k)$ 
is applied over $\ndel$ consecutive steps. 
The deviation bound given by \eqref{eq:deviation_bound} 
can then be conservatively adjusted by 
replacing $T$ with $\ndel T$ and by scaling the Lipschitz constants accordingly. 
This yields a more conservative estimate, 
ensuring constraint satisfaction even under known, 
fixed delays without requiring explicit multi-step simulation. 
\end{remark}
% ---------------------------------------------------------------------------
\paragraph{Local safety enforcement:} 
To align the theoretical guarantee with box-type constraints, 
we introduce component-wise deviation checks:
\begin{equation}
\label{eq:component_check}
\norm{x_i(k + 1) - \xbarplus_i } \leq \delta_i(k)
\end{equation}
Here $\delta_i(k)$ denotes the predicted deviation bound for the $i$-th state component. 

At deployment, Lipschitz constants are estimated locally. Hence, we have: 
\begin{equation}
\label{eq:delta_expression}
\delta_i(k) = \left(1 + T \lxi{i} \right) \, |x_i(k) - \Bar{x}_i| + T \luloc \, \norm{\bu^*(k) - \ubar}
\end{equation}
where $\lxi{i}$ (component-wise, for the $i$-th state component) 
and $\luloc$ (without a bar) are \emph{local} Lipschitz constant 
computed at time step $k$. 
This formulation ensures that each state component remains within its 
admissible deviation based on local sensitivity and proximity to verified neighbors. 
If \eqref{eq:component_check} holds for 
all  $i \in \{1, \ldots, n \}$ and   
all selected neighbor states,  
the interpolated control input $\bu^*(k)$ is deemed safe. 
Otherwise, the controller triggers a fallback mechanism, 
e.g., applying a conservative default input or initiating local retraining, to maintain safety.%

This two-level approach (global bound for theoretical validity and local checks for practical enforcement) 
enables scalable, certifiable deployment of learned policies in safety-critical systems 
without requiring online optimization or multi-step prediction.%

% ---------------------------------------------------------------------------
\subsubsection{Deployment-Time Safety Condition}

Finally, we compare the predicted deviation against the available safety margin. 
For each neighbor $\xbar$ used in the interpolation, let the associated verified successor $\xbarplus$ be strictly safe. 
The available safety room for $\xbarplus$ is quantified via a margin 
$\barbdelta$, computed component-wise as the minimum distance to the boundary 
of the merged safe set $\Bar{\mathcal{X}}^{\text{safe}}$, i.e.:
\begin{subequations}
\begin{align}
\label{eq:safety_margin}
&\bardelta_i =  \inf_{\bm{\xi} \in \partial\Bar{\mathcal{X}}^{\text{safe}}} \norm{\xbarplus_i - \xi_i}, \qquad i = 1, \ldots, n\\
&\Bar{\mathcal{X}}^{\text{safe}} = \bigcap_{\repd \in \mathcal{W}} \xsafe{\xbar,\repd}
\end{align}    
\end{subequations}
% \begin{equation}
% \bardelta = \text{distance}\left(\xbar,\ \partial \mathcal{X}^{\text{safe}} \right)
% \end{equation}
where $\partial \Bar{\mathcal{X}}^{\text{safe}}$ denotes the 
boundary of the merged safe set. 
Safety is guaranteed if we have: 
\begin{equation}
\label{eq:safety_check_component}
\hspace{-5ex}
\delta_i(k) \leq \bardelta_i,\quad \forall i \in \{1, \ldots, n\},\ \forall \text{ selected neighbors}
\end{equation} 
with $\delta_i(k)$ computed by \eqref{eq:delta_expression}. 
If \eqref{eq:safety_check_component} holds for all selected neighbors of $\xdep(k)$, 
then $ \xdep(k+1)$ remains within the merged safe set $\Bar{\mathcal{X}}^{\text{safe}}$. 
Otherwise, the controller applies a fallback input or expands the database.% 

% --------------------------------------------------------------------------------------

% \subsubsection{Runtime Formulation}

% At runtime, the safety filter checks whether the interpolated control input 
% $\bu^*(k)$, computed via nearest-neighbor interpolation, yields a safe next-step state. 
% This is evaluated for each neighbor $\xbar$ involved in the interpolation. 
% For each tuple $ \left(\xbar, \ubar, \xbarplus\right) $, 
% the following deviation bound is computed:
% \begin{equation}
% \delta(k)^{(j)} = \left( 1 + T \lx \right) \norm{\xdep(k) - \xbar } 
% + T \lu \norm{\bu^*(k) - \ubar}
% \end{equation}

% This quantity bounds the deviation between the interpolated next state $ \bx_{\text{curr}+1}(\bu^*) $ and the verified safe outcome $ \bx_{k+1}^{(j)} $, based on differences in state and control. The bound is compared to the available safety margin:
% \begin{equation}
% \delta(k)^{(j)} < \left( d^{\text{safe}} \right)^{(j)}.
% \end{equation}

% If this inequality holds for all selected nearest neighbors $ j = 1, \dots, N_{\text{NN}} $ used in the interpolation, the input $ \bu^* $ is deemed safe and applied. If not, the controller expands the database with new safe rollouts from the current state. This ensures safety during deployment while supporting generalization to previously unseen states.

\subsubsection{Estimating Lipschitz Constants for Safety Bounds}

Deployment-time safety relies on Lipschitz constants that 
quantify the sensitivity of the dynamics 
$\bmf(\bx,\bu,\repd)$ to changes in state and control. 
These constants bound the deviation of the interpolated next-step 
state $\xdep(k + 1)$ from its nearest certified neighbor. 

% ---------------------------------------------------------------
\paragraph{Control sensitivity:}  
The scalar constant $ \luloc $ captures the sensitivity to control perturbations, i.e.:
\begin{align}
\label{eq:Lu}
\luloc \approx
\sup_{\Delta \bu(k)} \frac{ \wnorm{ \bm{f}\left(\xdep(k), \bu^*(k) + \Delta \bu(k), \repd \right) 
- \bmf\left(\xdep(k), \bu^*(k),  \repd \right) }}{ \norm{\Delta \bu(k) }} 
\end{align}
where $\wnorm{\cdot}$ denotes a weighted Euclidean norm, i.e.:
\begin{equation}
\label{eq:weighted_norm}
\|z\|_{W} = \sqrt{z^\top W z}, \qquad 
W = \text{diag}\!\left(\frac{1}{c_1^2},\dots,\frac{1}{c_n^2}\right),
\end{equation}
with $c_1,\dots,c_n$ normalizing typical magnitudes of each state component.%

% ---------------------------------------------------------------
\paragraph{State sensitivity:}  

For each state component $i$, the local Lipschitz constant is defined as: 
\begin{align}
\label{eq:local_lx}
\lxi{i} \approx
\sup_{\Delta x_i(k)} 
\frac{\wnorm{\bmf\left(\xdep(k)+\Delta x_i(k)\unitvec_i,\bu^*(k) , \repd \right) 
- \bmf\left(\xdep(k),\bu^*(k), \repd \right)}}
{|\Delta x_i(k)|} 
\end{align}
where $\unitvec_i$ is the unit vector in the $i$-th direction. 

In practice, the supremum in \eqref{eq:local_lx} is approximated by 
evaluating the dynamics under a very small perturbation of the $i$-th state component. 
This yields a local estimate of $\lxi{i}$ without requiring exhaustive search.
The vector $L_{\bx}$ of all local state Lipschitz constants is 
$
L_{\bx} = \left[\lxi{1},\dots, \lxi{n}\right]^\top
$, which is recomputed online every step for local validity.%

\begin{remark}{Remark on Robustness:} 
While local Lipschitz constants provide practical safety bounds, 
their estimation introduces conservatism. 
To mitigate this, adaptive scaling or interval analysis may be employed to refine bounds under varying operating conditions. Future work will explore global Lipschitz estimation for stronger guarantees.
\end{remark}
% ------------------------------------------------------------
% ------------------------------------------------------------

% \subsubsection{Summary and Conclusions}

% This section presented a Lipschitz-based safety filter for runtime deployment of learned controllers. By interpolating control actions from a local database of verified safe transitions, and bounding deviations using locally estimated Lipschitz constants $ L_u $ and $ L_x $. The filter certifies that the resulting one-step state remains within the safe set.

% The method is lightweight, modular, and simulation-free, in contrast to many rollout-based safety filters that require online forward simulation to predict outcomes. By relying solely on local state-control data and deviation bounds, the approach enables safety generalization to previously unseen states without incurring simulation overhead. The formulation can also be extended to account for known actuation delays by adjusting the deviation bound over multiple timesteps. This allows the filter to maintain guarantees even when control inputs are applied with a fixed delay. If a violation is detected, the controller triggers database expansion to recover coverage. This enables reliable policy deployment with formal safety guarantees and minimal runtime overhead in nonlinear systems.

% -----------------------------------------------------------------
%      CASE STUDY & RESULTS
% -----------------------------------------------------------------

\section{Case Study: Aeroelastic Wing Control}
\label{sec:case_study}

To demonstrate the effectiveness of the proposed \mpcrl\ framework in a realistic safety-critical setting, the control of an aeroelastic wing subject to turbulence is considered in this case study. 
This system exemplifies the challenges addressed in this work, i.e., 
nonlinear dynamics, actuator delays, and strict state and input constraints.%

% -----------------------------------------------------------------

\subsection{System Setup}

The aeroelastic wing model includes five states: plunge $\pl$, 
pitch angle $\pitch$, plunge velocities $\vpl$, 
pitch velocity $\vpitch$, and flap deflection angle $\Dfl$. Each simulation begins from a random initial state within the operational envelope.
The control input is the commanded flap deflection angle $u(k) = \phi$, 
while disturbances are vertical turbulence generated using a Dryden model. 
The controller operates at $1$~kHz with an $8$~ms actuation delay. 
State and input constraints ensure structural integrity and actuation feasibility. 
In the presence of vertical turbulence $w^{\text{c}}(t)$, the effective aeroelastic angle of attack is defined as
\begin{align}
\label{eq:alpha_eff}
\alpha^{\text{eff,c}}(t) = 
\tan^{-1}\left( \frac{V \sin(\theta^{\text{c}}(t)) - w^{\text{c}}(t)}{V \cos(\theta^{\text{c}}(t))} \right) 
+ \frac{ v^{\text{h,c}}(t)}{V} + a^\text{offset,c} b^\text{c}(t) \cdot \frac{v^{\uptheta,\text{c}}(t)}{V}
\end{align}
where $V$ denotes the free-stream airspeed, $a^\text{offset,c}$ is aerodynamic center offset coefficient, and $b^\text{c}(t)$ is the 
continuous-time wing chord. 
% -----------------------------------------------------------------

\subsection{Controller Design} 

During training, an \gls{lpv} element is used to compute 
constraint-satisfying bounds for each state–disturbance pair. 
This guides the \gls{rl} component to learn a safe policy within these limits, 
using a quadratic reward that penalizes state deviation and control effort. 
At deployment, the learned policy is combined with a Lipschitz-based safety filter, 
as explained in Section~\ref{sec:deployment_time_guarantees},  
to certify interpolated actions without online optimization.

% -----------------------------------------------------------------

\subsection{Evaluation Protocol}
\glsreset{aoa}

Performance is assessed over $10$~s simulations with a $5$~s turbulence phase followed by recovery. 
A Monte Carlo campaign of $1000$ runs with randomized gust profiles evaluates robustness. 
Metrics include:
\begin{itemize}
    \item 
    maximum overshoot in plunge and the effective aeroelastic angle of attack (denoted by $\OSpl$ and $\OSaoa$, respectively),
    \item 
    settling time for plunge and the effective aeroelastic angle of attack (denoted by $\STpl$ and $\STaoa$, respectively),
    \item 
    \gls{rms} velocities and actuator smoothness.
\end{itemize}

\begin{remark}   
The effective aeroelastic angle of attack $\alpha_{\text{eff}}(t)$ is a key aerodynamic parameter 
that governs lift generation and stall margins. 
In this case study, $\alpha_{\text{eff}}(t)$ serves as a critical performance metric, 
as it reflects how effectively the controller mitigates gust-induced aerodynamic loads. 
While $\alpha_{\text{eff}}(t)$ is not a direct state variable, it is computed from Eq.~\eqref{eq:alpha_eff}, 
thereby linking aerodynamic performance to the controlled states.
\end{remark}
% -----------------------------------------------------------------

\subsection{Results Summary}

Table~\ref{tab:metrics} compares the proposed \mpcrl\ framework 
with standalone \gls{lpv} and \gls{rl} with regards to time-domain 
performance metrics (maximum overshoot and settling time). 
\mpcrl\ achieves the lowest overshoot and best plunge stability, 
while \gls{lpv} converges fastest post-gust. 
Pure \gls{rl} reacts aggressively, but lacks smoothness, saturating actuator limits.

\begin{center}
    \captionsetup{type=table}
    \begin{tabular}{@{}l|cccc@{}}
        \toprule
        Controller &
        $\OSpl$ [m] & $\STpl$ [s] &
        $\OSaoa$ [deg] & $\STaoa$ [s] \\
        \midrule
        \gls{lpv}      & 0.0080 & 2.47 & 2.280 & 0.83 \\
        \gls{rl}           & 0.0081 & 3.77 & 2.184 & 2.73 \\
        \mpcrl\       & \textbf{0.0069} & \textbf{2.02} & \textbf{1.726} & 0.98 \\
        \bottomrule
    \end{tabular}
    \captionof{table}{Time-domain performance metrics (mean over 1000 runs): 
    Metrics include overshoot and settling time for plunge and the effective aeroelastic angle of attack, reflecting structural and aerodynamic stability.}
    \label{tab:metrics}
\end{center}

To assess response smoothness and actuator efficiency, Table~\ref{tab:rms} reports RMS velocities and median actuator increments. These metrics are relevant because smoother responses reduce structural fatigue and improve long-term reliability.

\begin{table}[h!]
\centering
\caption{\gls{rms} velocities and actuator activity (mean over 1000 runs): 
\gls{rms} values in m/s and rad/s; actuator increments as \% of max step.}
\label{tab:rms}
\begin{tabular}{lccc}
\toprule
Controller & RMS $v^{\text{h}}$ [m/s] & RMS $v^{\upalpha}$ [rad/s] & Median $\Delta u$ [\%] \\
\midrule
\gls{lpv}    & 0.0422 & 0.1699 & 0.62 \\
\gls{rl}         & 0.0398 & 0.1614 & 100.0 \\
\mpcrl\    & \textbf{0.0348} & \textbf{0.1422} & 3.43 \\
\bottomrule
\end{tabular}
\end{table}

% -------------------------------------------------------------------------

\begin{figure}
\vspace{-5ex}
  \centering
  \begin{subfigure}{0.6\linewidth}
    \centering
    \includegraphics[width=\linewidth]{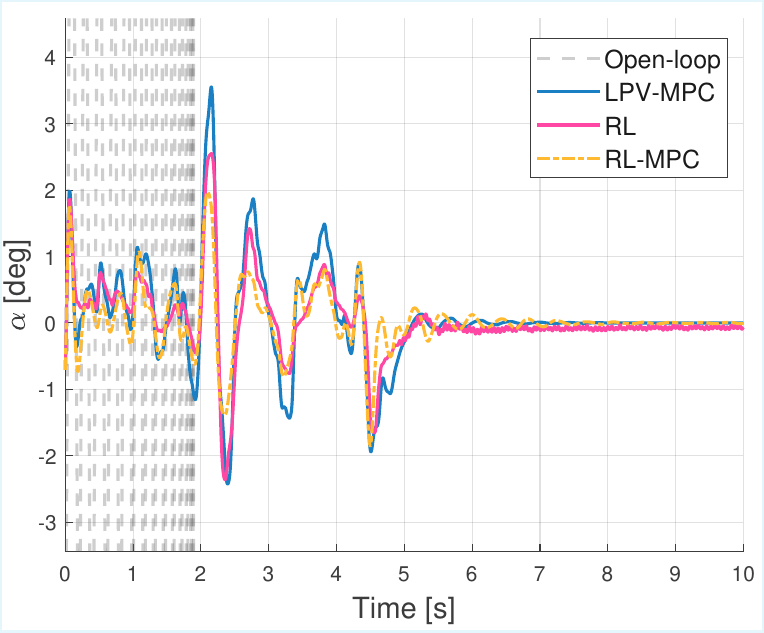}
    \caption{Effective aeroelastic angle of attack}
    \label{fig:responses-alpha}
  \end{subfigure}

  \begin{subfigure}{0.6\linewidth}
    \centering
    \includegraphics[width=\linewidth]{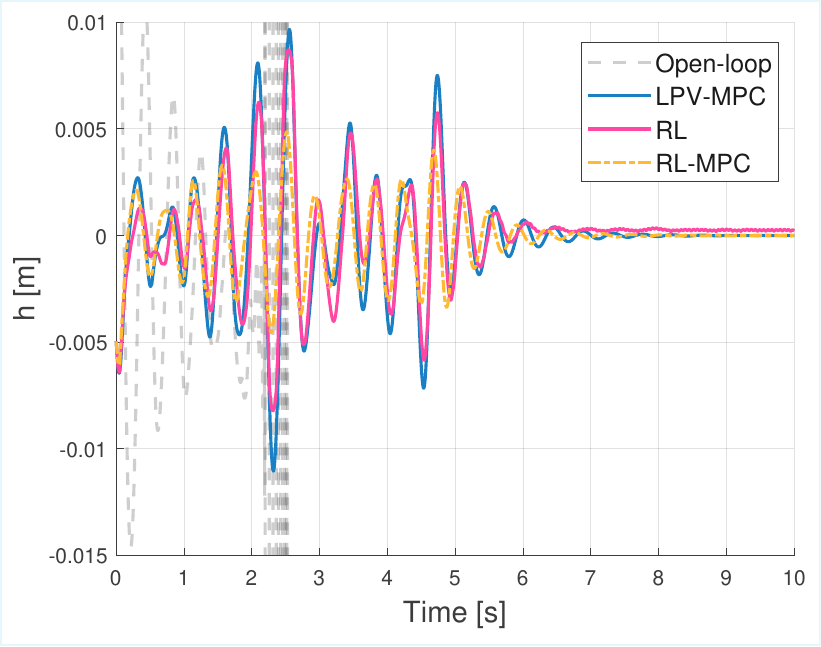}
    \caption{Plunge}
    \label{fig:responses-plunge}
  \end{subfigure}

  \begin{subfigure}{0.6\linewidth}
    \centering
    \includegraphics[width=\linewidth]{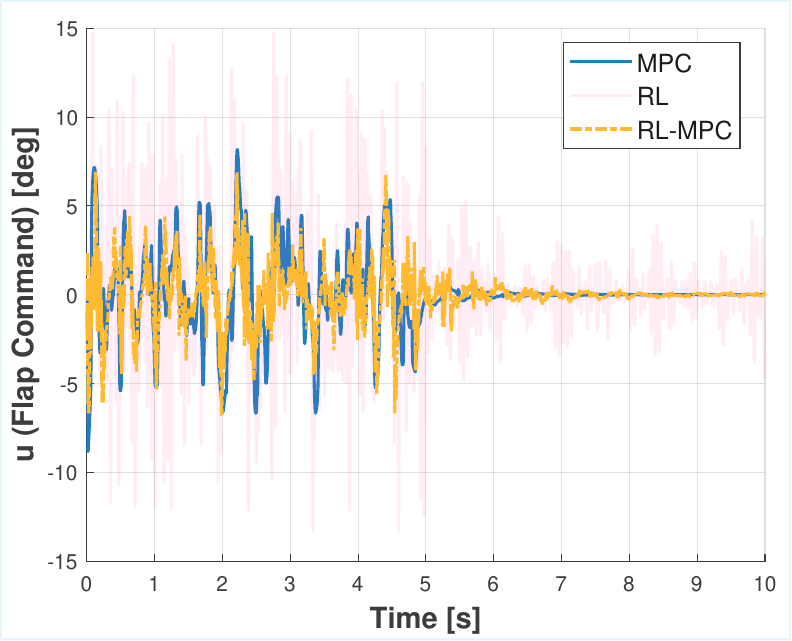}
    \caption{Flap command}
    \label{fig:responses-flap}
  \end{subfigure}

  \caption{Representative responses under gust and their corresponding control input (flap command).}
  \label{fig:responses}
\end{figure}

Figure~\ref{fig:responses} illustrates representative time-domain responses for angle of attack, plunge, and control input under identical gust conditions. RL–MPC offers the best trade-off between disturbance rejection and actuator smoothness.

\subsection{Core Takeaway}
The RL–MPC architecture combines constraint handling capabilities of \gls{mpc} 
with adaptability of \gls{rl}. 
This way it delivers superior gust-phase performance and robust safety guarantees. 

While validated on an aeroelastic wing, the approach generalizes to other nonlinear, 
safety-critical systems, e.g., robotic manipulators and autonomous vehicles.

\section{Conclusions \& Topics for Future Research}
\label{sec:paper_conclusion}
\glsresetall

Modern control systems should satisfy strict input and state constraints, 
reject unpredictable disturbances, and operate in real time despite nonlinear dynamics and limited onboard computation. 
Traditional \gls{mpc} offers structured constraint handling, 
but suffers under model mismatch and computational burden, 
while \gls{rl} provides adaptability but lacks safety guarantees.

This paper introduced an integrated \mpcrl\ architecture that combines predictive structure of \gls{mpc} with adaptability of \gls{rl}. 
Validated on a nonlinear aeroelastic wing, the approach achieves robust disturbance rejection and constraint satisfaction. 
Among tested controllers, \mpcrl\ delivered the smallest overshoot, 
tightest plunge tracking, and strongest low-frequency error attenuation, 
while maintaining actuator smoothness closer to \gls{lpv} than pure \gls{rl}. 
\gls{lpv} converged fastest post-gust, highlighting the value of exact models in calm conditions, but \mpcrl\ offered the best overall trade-off across metrics.

The framework generalizes beyond aeroelastic wings to other nonlinear, safety-critical systems, such as morphing aircraft, robotics, and automotive platforms, 
where real-time computation is limited. 
Its core design of \emph{offline long-horizon constrained learning 
with lightweight online deployment} provides a practical route for 
applying high-performance, constraint-aware \gls{rl} controllers in domains where classical methods struggle.

\paragraph*{Limitations and Future Work:} 
Key extensions to this work include refining reward design to reduce residual oscillations, 
formally bounding higher-order terms for stronger stability guarantees, 
and ensuring convergence in addition to constraint satisfaction. 
Experimental validation on physical platforms 
(e.g., wind tunnel tests) and application to diverse systems will further validate robustness. 

Assumption~\ref{ass:monotone} on monotonic control influence may not hold globally; 
systems with strong nonlinearities or coupled inputs may require alternative safety certification methods, such as local reachability or robust invariant set analysis.

In summary, combining offline \gls{mpc} with online \gls{rl} 
yields a controller that is both constraint-aware and robust to nonlinear disturbances. 
This offers a scalable foundation for safe, high-performance control in resource-constrained, safety-critical environments.

\bibliographystyle{elsarticle-num} 
\bibliography{sample}

@article{sun2022event,
  title={Event-triggered intelligent critic control with input constraints applied to a nonlinear aeroelastic system},
  author={Sun, Bo and Wang, Xuerui and Van Kampen, Erik-Jan},
  journal={Aerospace Science and Technology},
  volume={120},
  pages={107279},
  year={2022},
  publisher={Elsevier}
}

@article{wang2019stability,
  title={Stability analysis for incremental nonlinear dynamic inversion control},
  author={Wang, Xuerui and van Kampen, Erik-Jan and Chu, Qiping and Lu, Peng},
  journal={Journal of Guidance, Control, and Dynamics},
  volume={42},
  number={5},
  pages={1116--1129},
  year={2019},
  publisher={American Institute of Aeronautics and Astronautics},
  url={https://doi.org/10.2514/6.2018-1115},
  doi={10.2514/6.2018-1115}
}

@book{camacho2004mpcbook,
  title={Model Predictive Control},
  author={Camacho, Eduardo F. and Bordons Alba, Carlos},
  edition={2nd},
  publisher={Springer},
  year={2004},
  url={https://link.springer.com/book/10.1007/978-3-319-24853-0},
  doi={10.1007/978-3-319-24853-0}
}

@inproceedings{D-MPC-MLA,
  author    = {Mateus D. Virgilio Pereira and Ilya Kolmanovsky and Carlos E. Cesnik and Fabio Vetrano},
  title     = {Model Predictive Control Architectures for Maneuver Load Alleviation in Very Flexible Aircraft},
  booktitle = {AIAA Scitech 2019 Forum},
  year      = {2019},
  pages     = {1591},
  publisher = {American Institute of Aeronautics and Astronautics (AIAA)},
  doi       = {10.2514/6.2019-1591},
  url       = {https://doi.org/10.2514/6.2019-1591}
}

@inproceedings{D-MPC-LIDARpred,
  author    = {R. Bittner and H. Joos and T. Lombaerts and Q. Chu},
  title     = {Model Predictive Control for Maneuver Load Alleviation},
  booktitle = {IFAC Proceedings Volumes},
  year      = {2012},
  volume    = {45},
  pages     = {199-204},
  doi       = {10.3182/20120823-5-NL-3013.00049},
  url       = {https://doi.org/10.3182/20120823-5-NL-3013.00049}
}

@inproceedings{D-MPC-LPVhe-su,
  author    = {Tianyi He and Weihua Su},
  title     = {Gust Alleviation of Highly Flexible Aircraft with Model Predictive Control},
  booktitle = {AIAA SCITECH 2023 Forum},
  year      = {2023},
  pages     = {586},
  publisher = {American Institute of Aeronautics and Astronautics (AIAA)},
  doi       = {10.2514/6.2023-0586},
  url       = {https://doi.org/10.2514/6.2023-0586}
}

@article{D-MPC-LPV1,
  author = {Tianyi He and Weihua Su},
  title = {Robust control of gust-induced vibration of highly flexible aircraft},
  journal = {Aerospace Science and Technology},
  volume = {143},
  year = {2023},
  url = {https://www.sciencedirect.com/science/article/abs/pii/S1270963823005990},
  doi = {10.1016/j.ast.2023.108703}
}

@article{D-MPC-LPV2,
  title={Gust load alleviation of a flexible flying wing with linear parameter-varying modeling and model predictive control},
  author={Gao, Wei and Liu, Yishu and Li, Qifu and Lu, Bei},
  journal={Aerospace Science and Technology},
  volume={155},
  pages={109671},
  year={2024},
  publisher={Elsevier},
  url = {https://doi.org/10.1016/j.ast.2024.109671},
  doi={10.1016/j.ast.2024.109671}
}

@inproceedings{darwich2022gust,
  title={Gust and Manoeuvre Loads Alleviation Using Lower Surface Spoiler},
  author={Darwich, Abdel},
  booktitle={AIAA SCITECH 2022 Forum},
  pages={0245},
  year={2022}
}

@article{D-MPC-wang,
  author = {Yinan Wang and Andrew Wynn and Rafael Palacios},
  title = {Model-Predictive Control of Flexible Aircraft Dynamics using Nonlinear Reduced-Order Models},
  journal = {AIAA SciTech},
  year = {2016},
  url = {https://doi.org/10.2514/6.2016-0711},
  doi = {10.2514/6.2016-0711}
}

@article{D-MPC-MaciejowskiEtAl-03,
  author    = {Jan M. Maciejowski and Colin N. Jones},
  title     = {{MPC} Fault-Tolerant Flight Control Case Study: {Flight} 1862},
  journal   = {IFAC Proceedings Volumes},
  volume    = {36},
  number    = {5},
  pages     = {119-124},
  year      = {2003},
  month     = {June},
  doi       = {10.1016/S1474-6670(17)36480-7},
  url       = {https://doi.org/10.1016/S1474-6670(17)36480-7}
}

@inproceedings{D-MPC-PereiraEtAl-19,
  author    = {Mateus de F. V. Pereira and Ilya Kolmanovsky and Carlos E. S. Cesnik},
  title     = {Model Predictive Control with Constraint Aggregation Applied to Conventional and Very Flexible Aircraft},
  booktitle = {2019 IEEE 58th Conference on Decision and Control (CDC)},
  year      = {2019},
  pages     = {431-437},
  publisher = {IEEE},
  doi       = {10.1109/CDC40024.2019.9029769},
  url       = {https://doi.org/10.1109/CDC40024.2019.9029769}
}

@incollection{deepRL-book,
  title={Introduction to Reinforcement Learning},
  author={Ding, Zihan and Huang, Yanhua and Yuan, Hang and Dong, Hao},
  editor={Dong, Hao and Ding, Zihan and Zhang, Shanghang},
  booktitle={Deep Reinforcement Learning: Fundamentals, Research, and Applications},
  publisher={Springer Nature},
  year={2020},
  chapter={2},
  pages={47--124},
  url={http://www.deepreinforcementlearningbook.org}
}

@article{RL-stabilization-2,
  author    = {Mohsen Zahmatkesh and Seyyed Ali Emami and Afshin Banazadeh and Paolo Castaldi},
  title     = {Attitude Control of Highly Maneuverable Aircraft Using an Improved Q-learning},
  journal   = {arXiv preprint arXiv:2210.12317},
  year      = {2022},
  url       = {https://doi.org/10.48550/arXiv.2210.12317}
}

@article{RL-adaptive-maneuvering-2,
  author    = {Ramesh Konatala and Daniel Milz and Christian Weiser and Gertjan Looye and Erik-Jan van Kampen},
  title     = {Flight Testing Reinforcement-Learning-Based Online Adaptive Flight Control Laws on {CS-25-Class} Aircraft},
  journal   = {Journal of Guidance, Control, and Dynamics},
  year      = {2024},
  volume    = {47},
  number    = {11},
  pages     = {},
  month     = {November},
  doi       = {10.2514/1.G008321},
  url       = {https://doi.org/10.2514/1.G008321}
}

@inproceedings{RL-fault-tolerant-1,
  author    = {Killian Dally and Erik-Jan van Kampen},
  title     = {Soft Actor-Critic Deep Reinforcement Learning for Fault Tolerant Flight Control},
  booktitle = {AIAA SCITECH 2022 Forum},
  year      = {2022},
  pages     = {2078},
  publisher = {American Institute of Aeronautics and Astronautics (AIAA)},
  doi       = {10.2514/6.2022-2078},
  url       = {https://doi.org/10.2514/6.2022-2078},
}

@article{RL-GLA-1,
  title={Learning-based vs Model-free Adaptive Control of a MAV under Wind Gust},
  author={Chaffre, Thomas and Moras, Julien and Chan-Hon-Tong, Adrien and Marzat, Julien and Sammut, Karl and Le Chenadec, Gilles and Clement, Benoit},
  journal={arXiv preprint arXiv:2101.12501},
  year={2021},
  url = {https://doi.org/10.48550/arXiv.2101.12501},
  doi={10.48550/arXiv.2101.12501}
}

@article{RL-GLA-2,
  title={Deep reinforcement learning reveals fewer sensors are needed for autonomous gust alleviation},
  author={Haughn, Kevin PT and Harvey, Christina and Inman, Daniel},
  journal={arXiv preprint arXiv:2304.03133},
  year={2023},
  url = {https://doi.org/10.48550/arXiv.2304.03133},
  doi={10.48550/arXiv.2304.03133}
}

@article{RL-SYSID-3,
  author    = {Nathan Schaff},
  title     = {Online Aircraft System Identification Using a Novel Parameter Informed Reinforcement Learning Method},
  journal   = {Engineering Applications of Artificial Intelligence},
  year      = {2023},
  institution = {Embry-Riddle Aeronautical University},
  type      = {Master's thesis},
  url       = {https://commons.erau.edu/edt/779/},
}

@article{SafeRL+Robust-MPC,
  author = {Mario Zanon and Sébastien Gros},
  title = {Safe Reinforcement Learning Using Robust {MPC}},
  journal = {IEEE Transactions on Automatic Control},
  year = {2021},
  volume = {},
  number = {},
  pages = {},
  doi = {10.1109/TAC.2020.3024161},
  url = {https://doi.org/10.1109/TAC.2020.3024161},
  note = {Originally submitted to arXiv as 1906.04005v2 on 10 Jun 2019, last revised 17 Aug 2020},
  eprint = {1906.04005},
  archivePrefix = {arXiv},
  primaryClass = {eess.SY}
}

@article{RL+MPC-microgrids,
  title = {Integrating Reinforcement Learning and Model Predictive Control with Applications to Microgrids},
  author = {Caio Fabio Oliveira da Silva and Azita Dabiri and Bart De Schutter},
  journal = {arXiv preprint},
  volume = {arXiv:2409.11267},
  year = {2024},
  url = {https://doi.org/10.48550/arXiv.2409.11267},
  eprint = {2409.11267},
  archivePrefix = {arXiv},
  primaryClass = {eess.SY}
}

@inproceedings{RL-setpoints-MPC-control,
  title={A Hybrid Reinforcement Learning-{MPC} Approach for Distribution System Critical Load Restoration},
  author={Abinet Tesfaye Eseye and Xiangyu Zhang and Bernard Knueven and Matthew Reynolds and Weijia Liu and Wesley Jones},
  booktitle={2022 IEEE Power \& Energy Society General Meeting (PESGM)},
  year={2022},
  pages={1-5},
  doi={10.1109/PESGM48719.2022.9916743},
  publisher={IEEE},
  location={Denver, CO, USA},
  url={https://doi.org/10.1109/PESGM48719.2022.9916743}
}

@article{RL-guidance-MPC-lowlevel,
  author = {Greatwood, Colin and Richards, Arthur},
  year = {2019},
  month = {10},
  pages = {},
  title = {Reinforcement learning and model predictive control for robust embedded quadrotor guidance and control},
  volume = {43},
  journal = {Autonomous Robots},
  url={https://doi.org/10.1007/s10514-019-09829-4},
  doi = {10.1007/s10514-019-09829-4}
}

@article{MPC+DRL-TRAFFIC,
  author = {Dingshan Sun and Anahita Jamshidnejad and Bart De Schutter},
  title = {A Novel Framework Combining {MPC} and Deep Reinforcement Learning With Application to Freeway Traffic Control},
  journal = {IEEE Transactions on Intelligent Transportation Systems},
  volume = {25},
  number = {7},
  pages = {6756--6769},
  year = {2024},
  url = {https://doi.org/10.1109/TITS.2023.3342651},
  doi = {10.1109/TITS.2023.3342651},
  publisher = {IEEE},
  note = {Published under a Creative Commons License}
}

@inproceedings{MPC-+-RL-for-uncertainties-or-disturbances,
  author = {Willemijn Remmerswaal and Dingshan Sun and Anahita Jamshidnejad and Bart De Schutter},
  title = {Combined {MPC} and Reinforcement Learning for Traffic Signal Control in Urban Traffic Networks},
  booktitle = {Proceedings of the 26th International Conference on System Theory, Control and Computing, ICSTCC 2022},
  pages = {432--439},
  year = {2022},
  publisher = {IEEE},
  url = {https://doi.org/10.1109/ICSTCC55426.2022.9931771},
  doi = {10.1109/ICSTCC55426.2022.9931771}
}

@INPROCEEDINGS{MPC-general-RL-refines,
  author={Bang, Seung Hyeon and Jové, Carlos Arribalzaga and Sentis, Luis},
  booktitle={2024 IEEE-RAS 23rd International Conference on Humanoid Robots (Humanoids)}, 
  title={{RL-augmented MPC} Framework for Agile and Robust Bipedal Footstep Locomotion Planning and Control}, 
  year={2024},
  volume={},
  number={},
  pages={607--614},
  doi={10.1109/Humanoids58906.2024.10769914}
}

@article{MPC-general-RL-changes-offroad,
  author = {Prakhar Gupta and Jonathon M. Smereka and Yunyi Jia},
  title = {Reinforcement Learning Compensated Model Predictive Control for Off-road Driving on Unknown Deformable Terrain},
  journal = {arXiv:2408.09253 [cs.RO]},
  year = {2024},
  url = {https://doi.org/10.48550/arXiv.2408.09253},
  note = {Submitted on 17 Aug 2024}
}

@article{RL-based-on-RMPC,
  author    = {Sébastien Gros and Mario Zanon and Alberto Bemporad},
  title     = {Safe Reinforcement Learning via Projection on a Safe Set: How to Achieve Optimality?},
  journal   = {IFAC-PapersOnLine},
volume = {53},
number = {2},
pages = {8076-8081},
  year      = {2020},
  doi       = {10.48550/arXiv.2004.00915},
  url       = {https://www.sciencedirect.com/science/article/pii/S2405896320329360}
}

@inproceedings{RL-based-on-MPC+stoch-policy-gradient,
  author    = {Sébastien Gros and Mario Zanon},
  title     = {Reinforcement Learning based on {MPC} and the Stochastic Policy Gradient Method},
  booktitle = {2021 American Control Conference (ACC)},
  year      = {2021},
  pages     = {2590-2596},
  publisher = {IEEE},
  address   = {New Orleans, LA, USA},
  doi       = {10.23919/ACC50511.2021.9482765},
  url       = {https://doi.org/10.23919/ACC50511.2021.9482765},
  note      = {Conference Dates: 25-28 May 2021, Added to IEEE Xplore on 28 July 2021}
}

@article{Variance-exploration-for-RL-MPC,
  author    = {Arash Bahari Kordabad and Dirk Reinhardt and Akhil S Anand and Sébastien Gros},
  title     = {Reinforcement Learning for {MPC: Fundamentals} and Current Challenges},
  journal   = {IFAC-PapersOnLine},
  year      = {2023},
  volume    = {56},
  number    = {2},
  pages     = {5773--5780},
  doi       = {10.1016/j.ifacol.2023.10.548},
  url       = {https://doi.org/10.1016/j.ifacol.2023.10.548},
  note      = {Published in January 2023},
  publisher = {Elsevier}
}

@article{multi-agent-RL-via-DMPC,
  author    = {Samuel Mallick and Filippo Airaldi and Azita Dabiri and Bart De Schutter},
  title     = {Multi-Agent Reinforcement Learning via Distributed {MPC} as a Function Approximator},
  journal   = {arXiv preprint arXiv:2312.05166},
  year      = {2024},
  doi       = {10.48550/arXiv.2312.05166},
  url       = {https://doi.org/10.48550/arXiv.2312.05166},
  note      = {Accepted for publication in Automatica, arXiv:2312.05166v4 [eess.SY]},
  howpublished = {\url{https://arxiv.org/abs/2312.05166}},
  submissionhistory = {Submitted on 8 Dec 2023 (v1), last revised 18 Dec 2024 (v4)}
}

@article{overview-of-RL-for-MPC-fundamentals-and-challenges,
  author    = {Rudolf Reiter and Jasper Hoffmann and Dirk Reinhardt and Florian Messerer and Katrin Baumgärtner and Shamburaj Sawant and Joschka Boedecker and Moritz Diehl and Sébastien Gros},
  title     = {Synthesis of Model Predictive Control and Reinforcement Learning: Survey and Classification},
  journal   = {arXiv preprint arXiv:2502.02133},
  year      = {2025},
  doi       = {10.48550/arXiv.2502.02133},
  url       = {https://doi.org/10.48550/arXiv.2502.02133},
  note      = {arXiv:2502.02133v1 [eess.SY]},
  howpublished = {\url{https://arxiv.org/abs/2502.02133}},
  submissionhistory = {Submitted on 4 Feb 2025 (v1)}
}

@article{overview-of-MPC-+-RL-survey-and-classification,
  author    = {Katrine Seel and Alberto Bemporad and Sébastien Gros and Jan Tommy Gravdahl},
  title     = {Variance-Based Exploration for Learning Model Predictive Control},
  journal   = {IEEE Access},
  year      = {2023},
  volume    = {11},
  pages     = {60724--60736},
  doi       = {10.1109/ACCESS.2023.3282842},
  url       = {https://doi.org/10.1109/ACCESS.2023.3282842},
  publisher = {IEEE},
  note      = {Published under Creative Commons License},
  electronicISSN = {2169-3536}
}

@article{SYSID+RLMPC,
  author    = {Andreas B. Martinsen and Anastasios M. Lekkas and Sébastien Gros},
  title     = {Combining system identification with reinforcement learning-based {MPC}},
  journal   = {arXiv preprint arXiv:2004.03265},
  year      = {2020},
  doi       = {10.48550/arXiv.2004.03265},
  url       = {https://doi.org/10.48550/arXiv.2004.03265},
  note      = {Accepted to the IFAC 2020, arXiv:2004.03265v1 [eess.SY]},
  howpublished = {\url{https://arxiv.org/abs/2004.03265}},
  submissionhistory = {Submitted on 7 Apr 2020 (v1)}
}

@article{RL-based-on-real-time-NMPC,
  author    = {Mario Zanon and Vyacheslav Kungurtsev and Sébastien Gros},
  title     = {Reinforcement Learning Based on Real-Time Iteration {NMPC}},
  journal   = {arXiv preprint arXiv:2005.05225},
  year      = {2020},
  doi       = {10.48550/arXiv.2005.05225},
  url       = {https://doi.org/10.48550/arXiv.2005.05225},
  note      = {Accepted for the IFAC World Congress 2020, arXiv:2005.05225v1 [eess.SY]},
  howpublished = {\url{https://arxiv.org/abs/2005.05225}},
  submissionhistory = {Submitted on 11 May 2020 (v1)}
}

@article{safe-RL-using-NMPC-and-Policy-gradients,
  author = {Sébastien Gros and Mario Zanon},
  title = {Towards Safe Reinforcement Learning Using {NMPC} and Policy Gradients: {Part I - Stochastic} Case},
  journal = {arXiv preprint},
  volume = {arXiv:1906.04057},
  year = {2019},
  doi = {10.48550/arXiv.1906.04057},
  url = {https://doi.org/10.48550/arXiv.1906.04057},
  eprint = {1906.04057},
  archivePrefix = {arXiv},
  primaryClass = {eess.SY},
  note = {Submitted on 10 Jun 2019 (v1)}
}

@article{RL-to-tune-NMPC,
  author    = {Saket Adhau and Sébastien Gros and Sigurd Skogestad},
  title     = {Reinforcement learning based {MPC} with neural dynamical models},
  journal   = {European Journal of Control},
  year      = {2024},
  volume    = {101},
  article   = {101048},
  doi       = {10.1016/j.ejcon.2024.101048},
  url       = {https://doi.org/10.1016/j.ejcon.2024.101048},
  note      = {Open Access under Creative Commons license},
}

@inproceedings{RL-for-improving-MPC-params,
  author    = {Katrine Seel and Sébastien Gros and Jan Tommy Gravdahl},
  title     = {Combining Q-learning and Deterministic Policy Gradient for Learning-Based {MPC}},
  booktitle = {2023 62nd IEEE Conference on Decision and Control (CDC)},
  year      = {2023},
  pages     = {10383562},
  doi       = {10.1109/CDC49753.2023.10383562},
  publisher = {IEEE},
  address   = {Singapore, Singapore},
  url       = {https://doi.org/10.1109/CDC49753.2023.10383562},
}

@article{adaptive-parametrized-MPC-based-on-RL+synethis-framework,
  author    = {Dingshan Sun and Anahita Jamshidnejad and Bart De Schutter},
  title     = {Adaptive parameterized model predictive control based on reinforcement learning: A synthesis framework},
  journal   = {Engineering Applications of Artificial Intelligence},
  year      = {2024},
  volume    = {109},
  pages     = {109009},
  doi       = {10.1016/j.engappai.2024.109009},
  url       = {https://doi.org/10.1016/j.engappai.2024.109009},
  note      = {Published under a Creative Commons license},
}

@article{RL-within-MPC-framework,
  author    = {Filippo Airaldi and Bart De Schutter and Azita Dabiri},
  title     = {Reinforcement Learning with Model Predictive Control for Highway Ramp Metering},
  journal   = {arXiv preprint arXiv:2311.08820},
  year      = {2023},
  url       = {https://doi.org/10.48550/arXiv.2311.08820},
  note      = {Submitted to IEEE Transactions on Intelligent Transportation Systems},
}

@article{LQR1,
  author    = {William L. Garrard and Bradley S. Liebst},
  title     = {Active Flutter Suppression Using Eigenspace and Linear Quadratic Design Techniques},
  journal   = {Journal of Guidance, Control, and Dynamics},
  volume    = {8},
  number    = {3},
  pages     = {304--311},
  year      = {1985},
  publisher = {AIAA},
  doi       = {10.2514/3.19980},
  url       = {https://doi.org/10.2514/3.19980}
}

@article{LQR2,
  author    = {Labane Chrif and Zemalache Meguennni Kadda},
  title     = {Aircraft Control System Using {LQG and LQR} Controller with Optimal Estimation–Kalman Filter Design},
  journal   = {Procedia Engineering},
  volume    = {80},
  pages     = {245--257},
  year      = {2014},
  publisher = {Elsevier},
  doi       = {10.1016/j.proeng.2014.09.084},
  url       = {https://doi.org/10.1016/j.proeng.2014.09.084}
}

@book{LQRDISTURBACE,
  author    = {Brian D. O. Anderson and John B. Moore},
  title     = {Optimal Control: Linear Quadratic Methods},
  publisher = {Dover Publications},
  year      = {2014},
  note      = {[Edition unavailable]},
  url       = {https://www.perlego.com/book/1444508/optimal-control-linear-quadratic-methods-pdf}
}

@article{WINGFLEX1,
  author    = {Qinfeng Guo and Xi He and Zhuo Wang and Jinjun Wang},
  title     = {Effects of Wing Flexibility on Aerodynamic Performance of an Aircraft Model},
  journal   = {Chinese Journal of Aeronautics},
  volume    = {34},
  number    = {9},
  pages     = {133--142},
  year      = {2021},
  publisher = {Elsevier},
  doi       = {10.1016/j.cja.2021.01.012},
  url       = {https://doi.org/10.1016/j.cja.2021.01.012}
}

@article{WINGFLEX2,
  author    = {Tobias Franziskus Wunderlich and Sascha D\"ahne and Lars Reimer and Andreas Schuster},
  title     = {Global Aerostructural Design Optimization of More Flexible Wings for Commercial Aircraft},
  journal   = {Journal of Aircraft},
  volume    = {58},
  number    = {849},
  year      = {2021},
  publisher = {AIAA},
  doi       = {10.2514/1.C036301},
  url       = {https://doi.org/10.2514/1.C036301}
}

@techreport{FLUTTER1,
  author       = {Eli Livne},
  title        = {Aircraft Active Flutter Suppression: State of the Art and Technology Maturation Needs},
  institution  = {U.S. Department of Transportation, Federal Aviation Administration},
  number       = {DOT/FAA/TC-18/47},
  year         = {2019},
  address      = {William J. Hughes Technical Center, Atlantic City, NJ},
  url          = {https://arc.aiaa.org/doi/10.2514/1.C034442}
}

@inproceedings{Hinf,
  author    = {A.D.P. Schoon and S.T. Theodoulis},
  title     = {Review of {H{$\infty$}} Static Output Feedback Controller Synthesis Methods for Fighter Aircraft Control},
  booktitle = {Proceedings of the AIAA SCITECH 2025 Forum},
  year      = {2025},
  publisher = {AIAA},
  pages     = {25},
  doi       = {10.2514/6.2025-2241},
  url       = {https://doi.org/10.2514/6.2025-2241},
  event     = {AIAA SCITECH 2025 Forum, Orlando, United States},
  note      = {ISBN: 978-1-62410-723-8}
}

@inproceedings{Gainscheduling,
  author    = {Guilhem Puyou and Caroline Berard},
  title     = {Gain-Scheduled Flight Control Law for Flexible Aircraft: A Practical Approach},
  booktitle = {IFAC Proceedings Volumes},
  volume    = {40},
  number    = {7},
  pages     = {497--502},
  year      = {2007},
  doi       = {10.3182/20070625-5-FR-2916.00085},
  url       = {https://doi.org/10.3182/20070625-5-FR-2916.00085},
  note      = {17th IFAC Symposium on Automatic Control in Aerospace}
}

@INPROCEEDINGS{LPVanalysis-gainscheduling,
  author={Hjartarson, Arnar and Seiler, Peter and Balas, Gary J.},
  booktitle={2014 American Control Conference}, 
  title={{LPV} analysis of a gain scheduled control for an aeroelastic aircraft}, 
  year={2014},
  volume={},
  number={},
  pages={3778-3783},
  url={https://doi.org/10.1109/ACC.2014.6859301},
  doi={10.1109/ACC.2014.6859301}
}

@article{INDI,
  author    = {Agnes Steinert and Stefan Raab and Simon Hafner and Florian Holzapfel and Haichao Hong},
  title     = {From Fundamentals to Applications of Incremental Nonlinear Dynamic Inversion: {A} Survey on {INDI – Part I}},
  journal   = {Chinese Journal of Aeronautics},
  year      = {2025},
  note      = {Available online 25 April 2025, In Press, Journal Pre-proof},
  doi       = {10.1016/j.cja.2025.103553},
  url       = {https://doi.org/10.1016/j.cja.2025.103553}
}

@article{NDI,
  author    = {Rasmus Steffensen and Agnes Steinert and Ewoud Jan Jacob Smeur},
  title     = {Non-Linear Dynamic Inversion with Actuator Dynamics: An Incremental Control Perspective},
  journal   = {arXiv preprint arXiv:2201.09805},
  year      = {2022},
  note      = {Version 2, last revised 25 Nov 2022},
  url       = {https://arxiv.org/abs/2201.09805},
  doi       = {10.48550/arXiv.2201.09805}
}

@book{Backstepping,
  author    = {Lungu, Mihai},
  title     = {Backstepping Control Method in Aerospace Engineering},
  year      = {2022},
  month     = {January},
  pages     = {},
  publisher = {Academica Greifswald},
  url       = {https://www.researchgate.net/publication/356786069-Backstepping-control-method-in-aerospace-engineering},
  isbn      = {978-3-940237-54-5}
}

@article{robust-nonlinear-general-1,
  author    = {Luis F. Canaza Ccari and Walker Aguilar and Elvis Supo and Erasmo Sulla Espinoza and Yuri Silva Vidal and Nicolás Medina},
  title     = {Robust Finite-Time Adaptive Nonlinear Control System for an EOD Robotic Manipulator: Design, Implementation, and Experimental Validation},
  journal   = {IEEE Access},
  volume    = {12},
  pages     = {93859--93875},
  year      = {2024},
  url       = {https://ieeexplore.ieee.org/document/10587221},
  doi       = {10.1109/ACCESS.2024.3424463},
  issn      = {2169-3536},
  publisher = {IEEE}
}

@inproceedings{robust-nonlinear-general-2,
  author    = {Siqi Hu and Edwin Babaians and Mojtaba Karimi and Eckehard Steinbach},
  title     = {{NMPC-MP: Real-time} Nonlinear Model Predictive Control for Safe Motion Planning in Manipulator Teleoperation},
  booktitle = {Proceedings of the 2021 IEEE/RSJ International Conference on Intelligent Robots and Systems (IROS)},
  pages     = {},
  year      = {2021},
  url       = {https://ieeexplore.ieee.org/document/9636802},
  doi       = {10.1109/IROS51168.2021.9636802},
  publisher = {IEEE},
  address   = {Prague, Czech Republic}
}

@inproceedings{Sanches2024-LPVsolver,
  author    = {Vinícius M. Sanches and Marcelo M. Morato and Julio E. Normey-Rico and Olivier Sename},
  title     = {An Interior-Point {LPV MPC} Solver for Real-time Systems: {Experimental} Processor-in-the-loop Validation},
  booktitle = {Proceedings of the 6th IFAC Workshop on Linear Parameter Varying Systems (LPVS'25)},
  pages     = {},
  year      = {2025},
  url       = {https://hal.science/hal-05028039},
  publisher = {HAL Open Science},
  address   = {Porto, Portugal.}
}

@article{Lehmann2024PolicyGradients,
  author    = {M. Lehmann},
  title     = {The Definitive Guide to Policy Gradients in Deep Reinforcement Learning: Theory, Algorithms and Implementations},
  journal   = {arXiv preprint arXiv:2401.13662},
  year      = {2024},
  url       = {https://doi.org/10.48550/arXiv.2401.13662}
}

@article{Shim2025,
  author    = {Dongjun Shim and Daniel Liberzon},
  title     = {Stability of linear systems with slow and fast time variation and switching: General case},
  journal   = {Nonlinear Analysis: Hybrid Systems},
  volume    = {58},
  pages     = {101626},
  year      = {2025},
  doi       = {10.1016/j.nahs.2025.101626}
}

@article{Murali2026,
  author    = {Vishnu Murali and Ashutosh Trivedi and Majid Zamani},
  title     = {Closure certificates},
  journal   = {Nonlinear Analysis: Hybrid Systems},
  volume    = {59},
  pages     = {101630},
   year      = {2026},
  doi       = {10.1016/j.nahs.2025.101630}
}

@article{HE2024101436,
title = {Robust model-based predictive iterative learning control for systems with non-repetitive disturbances},
journal = {Nonlinear Analysis: Hybrid Systems},
volume = {51},
pages = {101436},
year = {2024},
doi = {10.1016/j.nahs.2023.101436},
url = {https://www.sciencedirect.com/science/article/pii/S1751570X23001073},
author = {Chao He and Junmin Li and Sanyang Liu and Jiaxian Wang},
}

\newpage
% --------------------------------------------------
% Appendices
% --------------------------------------------------
\appendix

\section{Frequently-used mathematical notations}
\label{appendix:notations}

\begin{table}
\centering
\caption{Table of Notations used throughout the paper.}
\label{tab:notation}
\renewcommand{\arraystretch}{1}
\begin{tabularx}{\linewidth}{@{}l X l @{}}
\toprule
\textbf{Symbol} & \textbf{Description} & \textbf{Domain} \\
\midrule
$\bx(k)$          & Discrete-time system state vector at time step $k$ & $\mathbb{R}^n$ \\
$\bu(k)$          & Discrete-time control input vector at time step $k$               & $\mathbb{R}^m$ \\
$\bd(k)$          & Discrete-time disturbance vector at time step $k$                 & $\mathbb{R}^q$ \\
$\mathcal{X},\;\mathcal{U},\;\mathcal{W}$     & Admissible sets for state, 
control input, and disturbance &  $\mathbb{R}^n,\;\mathbb{R}^m,\;\mathbb{R}^q$ \\
$\xsafe{\bx(k),\bd(k)}$ &  Safe state set verified under constraints for state-input pair $\bx(k)$, $\bu(k)$          & $\mathcal{X}$ \\
$\umin,\,\umax$ & Element-wise lower and upper bounds for control input (from \gls{mpc}) &  $\mathbb{R}^m$ \\
$T$             & Sampling time                                       & time units \\
$\ell$          & Actuation delay horizon (number of discrete steps)  & integer \\
$\bbeta(k)$      & Actuator state vector (first-order low-pass model)  & $\mathbb{R}^m$ \\
% $\lambda_j$     & Actuator gain for component $j$.                     & scalar \\
$\lu(k)$        & Local Lipschitz constant w.r.t. control input at time step $k$ & scalar \\
$\lx^{(i)}(k)$  & Local Lipschitz constant of dynamics w.r.t. the $i$-th state component at time step $k$ & scalar \\
$\delta_i(k)$   & Predicted deviation bound for the $i$-th state component at time step $k$ & scalar \\
$\bar{\delta}_i$& Safety margin for the $i$-th state component (distance to constraint boundary). & scalar \\
$\prl(\cdot)$ & Reinforcement Learning policy                      & $\mathcal{X}\mapsto \mathcal{U}$ \\
$\pi^{\text{MPC}}(\cdot)$ 
& \gls{mpc} policy (first input of optimal sequence)       &  $\mathcal{X} \mapsto \mathcal{U}$ \\
% $\bar{X}_{\text{safe}}$ & Merged safe set over bounded disturbances used in deployment checks. & subset of $X$ \\
\bottomrule
\end{tabularx}
\end{table}
% ---------------------------------------------------
\section{Aeroelastic Wing Model and Assumptions}
\label{appendix:wing_model}

The aeroelastic wing model has nonlinear stiffness, aerodynamic-structural coupling, and actuator dynamics. The parameters in the model are identical to those in Ref.~\cite{sun2022event}. The continuous-time state vector is:
\[
\bxc(t) = \left[h^{\text{c}}(t), \theta^{\text{c}}(t), v^{\text{h,c}}(t), v^{\theta,\text{c}}(t), \beta_f^{\text{c}}(t) \right]^\top,
\]
where $h^{\text{c}}(t)$ is continuous-time plunge, 
$\theta^{\text{c}}(t)$ is continuous-time pitch angle, 
and $\beta_f^{\text{c}}(t)$ is continuous time flap deflection angle. The dynamics is governed by:
\[
\dot{\bx}^{\text{c}}(t) = \bmf\left(\bxc(t), u^{\text{c}}(t), w^{\text{c}}(t) \right),
\]
with $u^{\text{c}}(t)$ the continuous-time flap command 
and $w^{\text{c}}(t)$ the continuous-time vertical turbulence.

This model is derived based on the following assumptions:
\begin{itemize}
    \item 
    Disturbances are bounded and slowly time-varying.
    \item 
    Actuator dynamics is modeled as a first-order low-pass filter.
    \item 
    Monotonicity of the control influence holds locally.
\end{itemize}
For controller design, the system is discretized using a second-order Taylor expansion with 
sampling time $T = 0.001$~s.

% --------------------------------------------------

\newpage
\section{LPV Approximation of Aeroelastic Wing Model}
\label{appendix:validation}

The \gls{lpvmodel} model is constructed by linearizing the nonlinear dynamics at each time step:
\[
\bx(k+1) = A \left(\pl,\pitch \right)\bx(k) + B u(k),
\]
Here, $A \left(\pl,\pitch \right)$ is the \gls{lpvmodel} 
state matrix whose entries depend on plunge and pitch states 
and captures stiffness and aerodynamic variations. 
$B$ is  the input matrix for flap actuation and is assumed to be constant.%

\begin{figure}
\centering
\includegraphics[width=0.75\linewidth]{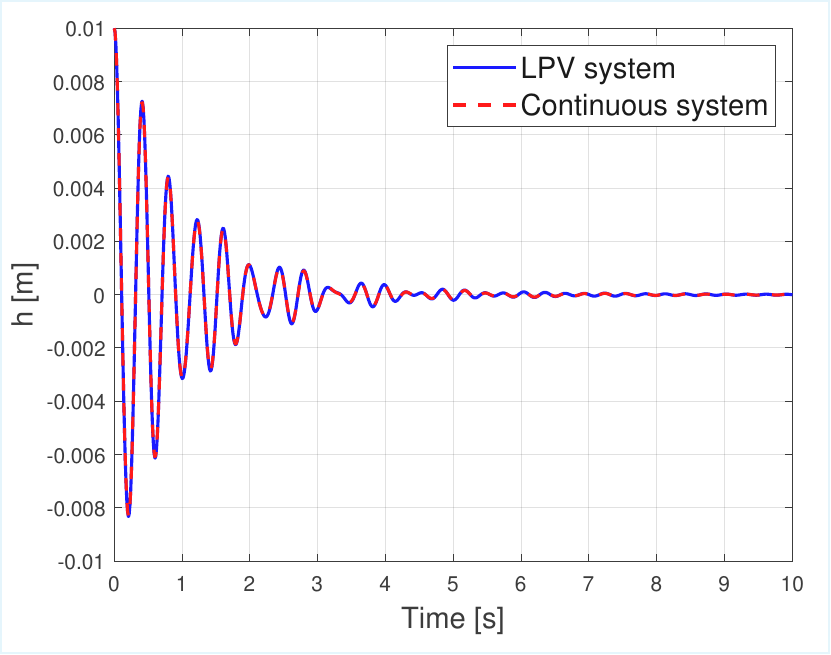}
\caption{Comparison of the \gls{lpvmodel} and continuous-time model responses.}
\label{fig:lpv-vs-continuous}

\includegraphics[width=0.75\textwidth]{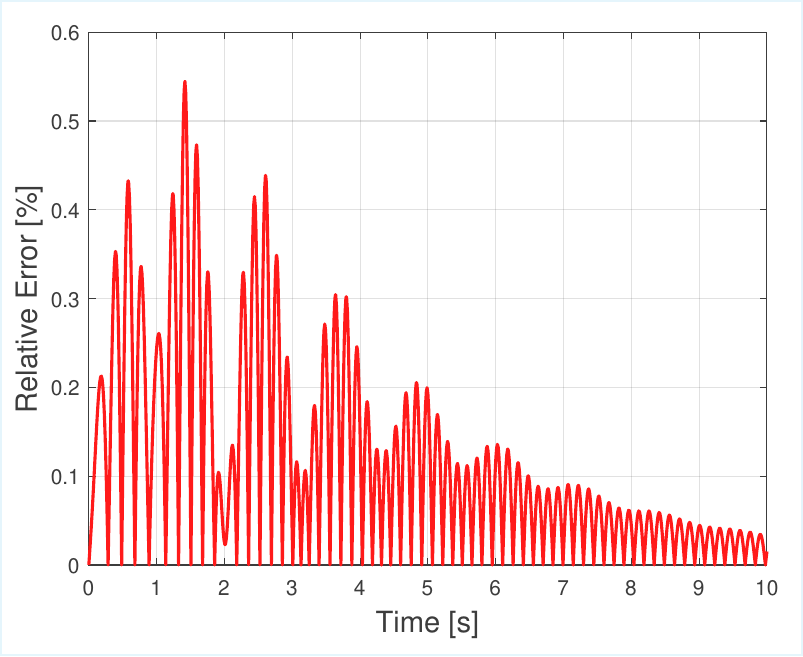}
\caption{The relative error of the  \gls{lpvmodel} model with respect to the 
continuous-time model.}
\label{fig:lpv-error}
\end{figure}

Validation of the \gls{lpvmodel} approximation confirms its suitability 
for the proposed control architecture. 
As shown in Figures~\ref{fig:lpv-vs-continuous} and \ref{fig:lpv-error}, 
the \gls{lpvmodel} model closely tracks the continuous nonlinear dynamics, 
with relative trajectory errors remaining below $0.156\%$. 
The maximum deviation in any state matrix entry is only $0.27\%$, 
which indicates that local linearizations provide an accurate representation 
of the underlying nonlinear behavior. 

Furthermore, a frequency-domain analysis reveals that the dominant 
system modes lie between $0.5$–$5$~Hz, well below the $1$~kHz 
update rate of the \gls{lpvmodel}-based controller. 
These results collectively demonstrate that the \gls{lpvmodel} 
approximation is both numerically precise and dynamically consistent, ensuring reliable constraint handling during offline MPC-based policy generation.%

% --------------------------------------------------
\newpage
\section{Extended Statistical Analysis}
\label{appendix:extended-results}

\begin{figure}[h!]
\centering
\includegraphics[width=0.8\linewidth]{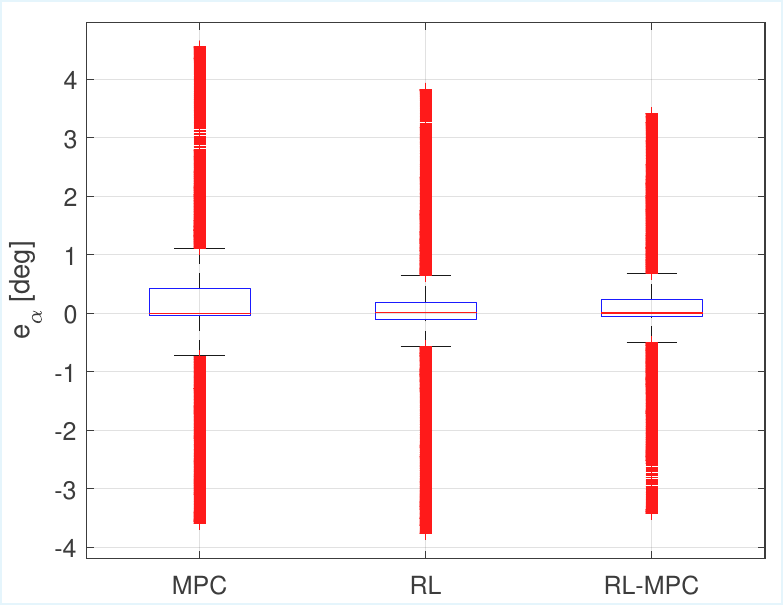}
\caption{The effective aeroelastic angle of attack error distribution (full run and post-gust).}
\label{fig:aoa-boxplot}

\includegraphics[width=0.8\linewidth]{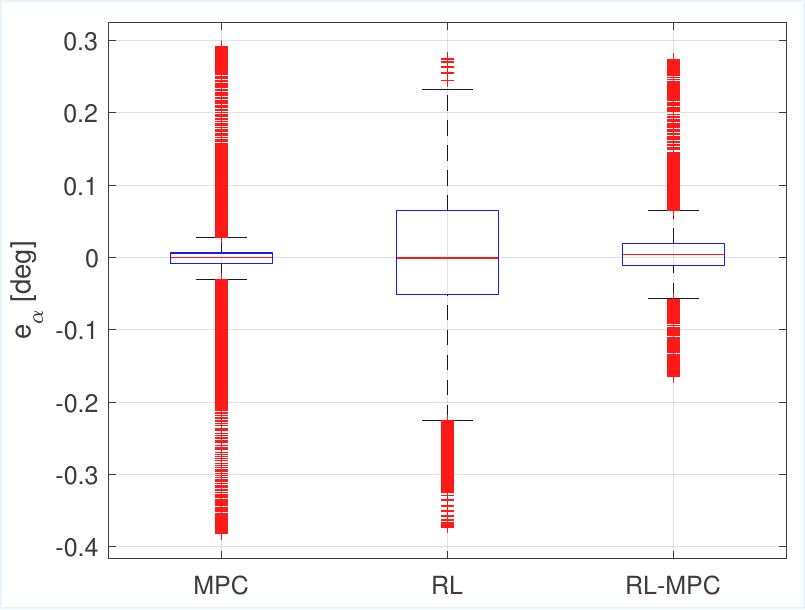}
\caption{The effective aeroelastic angle of attack error distribution (full run and post-gust).}
\label{fig:actuator-boxplot}
\end{figure}

Figures~\ref{fig:aoa-boxplot}–\ref{fig:actuator-boxplot} illustrate the distribution of the effective aeroelastic angle of attack errors and actuator activity under gust disturbances. 
Moreover, Tables~\ref{tab:aoa-metrics}–\ref{tab:freq-energy} summarize key statistical metrics, 
including excursions, \gls{rms} rates, and frequency-domain error energy across $1000$ Monte Carlo runs. 
Together, these results highlight differences in tracking accuracy, disturbance rejection, and control smoothness among the evaluated controllers.

These results reveal that the \mpcrl\ framework achieves 
the most consistent disturbance rejection, with the lowest error spread 
and fewest excursions, while \gls{lpv} excels in post-gust settling 
and actuator smoothness. 
Pure \gls{rl} reacts aggressively during gusts, 
but suffers from high actuator activity and slower recovery. 
Overall, \mpcrl\ offers the best trade-off between robustness, accuracy, and efficiency, 
combining strong gust-phase performance with acceptable control smoothness.%

\begin{table}[h!]
\centering
\caption{The effective aeroelastic angle of attack metrics: Excursions and \gls{rms} pitch rate.}
\label{tab:aoa-metrics}
\begin{tabular}{lccc}
\toprule
Controller & Average excursions &  \gls{rms} $v^{\upalpha}$ (full) & \gls{rms} $v^{\upalpha}$ (post-gust) \\
\midrule
\gls{lpv} & 5.20 & 0.1699 & 0.0108 \\
\gls{rl} & 3.54 & 0.1614 & 0.0291 \\
\mpcrl\ & \textbf{3.14} & \textbf{0.1422} & \textbf{0.0196} \\
\bottomrule
\end{tabular}

\vspace{10ex}

\caption{Plunge metrics over 1000 runs: Excursion threshold; $20\%$ of 
maximum open-loop gust response.}
\label{tab:freq-energy}
    \begin{tabular}{@{}lccc@{}}
        \toprule
        Controller &
        \makecell{Average excursions} &
        \makecell{\gls{rms} $\vpl$~$\left[\dfrac{\text{m}}{\text{s}}\right]$ (full)} &
        \makecell{\gls{rms} $\vpl$~$\left[\dfrac{\text{m}}{\text{s}}\right]$ (post-gust)} \\
        \midrule
        \gls{lpv}     & 4.7553  & 0.04216 & 0.0095 \\
        \gls{rl}          & 3.9991  & 0.03980 & 0.0099 \\
        \mpcrl\      & \textbf{3.0758}  & \textbf{0.03479} & \textbf{0.0092} \\
        \bottomrule
    \end{tabular}
\end{table}

% --------------------------------------------------

\end{document}